\documentclass[10pt,twocolumn,letterpaper]{article}

\usepackage[]{wacv} 

\usepackage{graphicx}
\usepackage{amsmath}
\usepackage{amssymb}
\usepackage{booktabs}
\usepackage{todonotes}
\usepackage{paralist}
\usepackage{multirow}
\usepackage{multicol}
\usepackage{subcaption} 
\usepackage{wrapfig}
\usepackage{xcolor}
\usepackage{makecell}
\usepackage{listings}
\usepackage{booktabs} 
\usepackage{enumitem} 
\usepackage{xcolor}
\usepackage[pagebackref,breaklinks,colorlinks]{hyperref}

\usepackage[capitalize]{cleveref}
\crefname{section}{Sec.}{Secs.}
\Crefname{section}{Section}{Sections}
\Crefname{table}{Table}{Tables}
\crefname{table}{Tab.}{Tabs.}

\def\wacvPaperID{2588} 
\def\confName{WACV}
\def\confYear{2025}

\begin{document}

\title{HiFi-CS: Towards Open Vocabulary Visual Grounding For \\ Robotic Grasping Using Vision-Language Models}

\author{Vineet Bhat, Prashanth Krishnamurthy, Ramesh Karri, Farshad Khorrami \\
New York University\\
Brooklyn, NY, USA\\
{\tt\small vrb9107@nyu.edu}
}
\maketitle

\begin{abstract}


Robots interacting with humans through natural language can unlock numerous applications such as Referring Grasp Synthesis (RGS). Given a text query, RGS determines a stable grasp pose to manipulate the referred object in the robot's workspace. RGS comprises two steps: visual grounding and grasp pose estimation. Recent studies leverage powerful Vision-Language Models (VLMs) for visually grounding free-flowing natural language in real-world robotic execution. However, comparisons in complex, cluttered environments with multiple instances of the same object are lacking. This paper introduces HiFi-CS, featuring hierarchical application of Featurewise Linear Modulation (FiLM) to fuse image and text embeddings, enhancing visual grounding for complex attribute rich text queries encountered in robotic grasping. Visual grounding associates an object in 2D/3D space with natural language input and is studied in two scenarios: Closed and Open Vocabulary. HiFi-CS features a lightweight decoder combined with a frozen VLM and outperforms competitive baselines in closed vocabulary settings while being 100x smaller in size. Our model can effectively guide open-set object detectors like GroundedSAM to enhance open-vocabulary performance. We validate our approach through real-world RGS experiments using a 7-DOF robotic arm, achieving 90.33\% visual grounding accuracy in 15 tabletop scenes. Our codebase is available at \url{https://github.com/vineet2104/hifics}.
\end{abstract}

\section{Introduction}
\label{sec:intro}

Language-guided robotic manipulation is crucial for the development of human-robot interactive systems. A key component of this is Referring Grasp Synthesis (RGS), which enables autonomous robots to execute pick-and-place tasks based on text commands. Given a request to grasp a specific object within its workspace, RGS identifies a stable grasp pose for execution using a robotic arm~\cite{tziafas2023language}. This process connects abstract natural language instructions with physical manipulation policies, forming a critical component of modern robotic visual perception \cite{9022259}. For instance, when given a command such as ``grasp the blue bottle," the RGS visual grounding module locates the referred ``blue bottle" in the robot's surroundings, either through 2D images \cite{liu2021refer,lu2023vl} or through a reconstructed 3D representation \cite{chen2020scanrefer,achlioptas2020referit_3d,zhao2021_3DVG_Transformer}. These visual representations are used to construct object point clouds, which are then fed into downstream grasping models to determine and execute the grasp pose \cite{gou2021RGB,fang2020graspnet,lu2022hybrid,fang2023anygrasp}.

\begin{figure}[!ht]
    \includegraphics[width=0.49\textwidth,trim=0in 0in 0in 0.0in,clip=true]{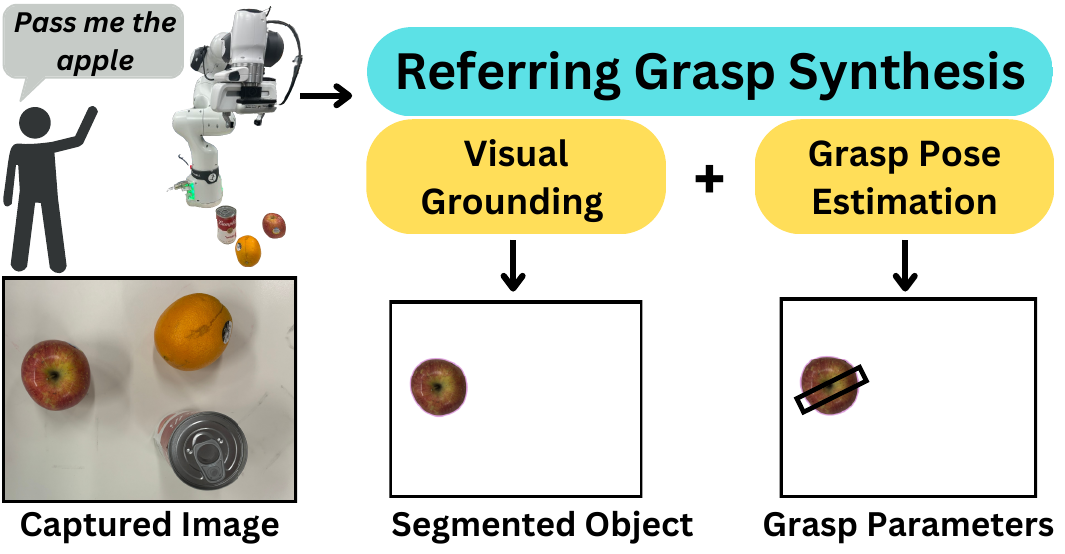}
  \caption{Referring Grasp Synthesis converts free-flowing language query to robot grasp pose.}
  \label{fig:rgs}
\end{figure}

The emergence of large-scale foundational models in both vision and language has further bridged the gap between robotic perception and real-world knowledge, offering promising advancements in RGS \cite{10342331}. These \textit{Vision-Language Models (VLMs)}, trained on vast datasets of real-world images and text, have demonstrated exceptional visual reasoning capabilities \cite{Radford2021LearningTV}. Consequently, VLMs have seen widespread adoption in RGS, which generally consists of two stages: Visual Grounding and Grasp Pose Estimation (\cref{fig:rgs}). Numerous works utilize VLMs for visual grounding, followed by pre-trained grasp detection modules \cite{lu2023vl,Mirjalili2023LANgraspUL,10161041,shen2023F3RM,lerftogo2023}. Some approaches study end-to-end RGS by directly training models to predict grasp poses from 2D/3D images \cite{tziafas2023language,Chen2021AJN,Jin2024ReasoningGV,Tang2023TaskOrientedGP}. An ideal RGS model should generate precise grasp poses for the target object in cluttered environments with multiple similar objects (distractors). The visual grounding stage must leverage object attributes such as color, shape, and relative position, as specified in text, to resolve ambiguity and demonstrate zero-shot capabilities in unseen environments. For robust performance, VG should handle simple queries like ``red apple" as well as complex ones such as ``grasp the red apple to the right of plate," when multiple similar objects are present.

In this paper, we frame RGS as a two-stage process. The first stage, Visual Grounding (VG), identifies the referred object in the captured image of the workspace based on the input natural language query. The second stage, Grasp Pose Estimation (GPE), determines the grasp parameters for the referred object. We present HiFi-ClipSeg (HiFi-CS), a language-conditioned 2D visual grounding model, and compare its performance with competitive baselines. HiFi-CS can accurately predict 2D segmentation masks from both simple and complex referring object queries in RGB images of the workspace, and its lightweight size allows fast fine-tuning and deployment capabilities ideal for robotic applications. Our contributions are as follows:

\begin{compactenum}

\item We propose a novel VG model that leverages a frozen VLM with a lightweight segmentation decoder. By applying hierarchical Featurewise Linear Modulation (FiLM) to fuse vision-text embeddings from the VLM, we enhance semantic retention, improving grounding of complex text queries in 2D space.
\item Our model surpasses existing methods in closed-vocabulary settings on two widely used robotic VG datasets, achieving an average Intersection over Union accuracy of 87\%. HiFi-CS outperforms open-set detectors like GroundedSAM by approximately 40\%. 
\item HiFi-CS can guide open-set object detectors, improving open-vocabulary performance on a new, challenging test dataset. Our RGS pipeline is deployed on a 7-DOF robotic arm in 15 real-world cluttered scenes, achieving a grounding accuracy of 90.33\%.
\end{compactenum}

\section{Background and Related Work}
\label{sec:rel_work
}

\noindent \textbf{Foundational Models in Robotics: } Large Language Models (LLMs) can generate high-level robotic execution plans based on task inputs and environmental context \cite{huang2022language, Song_2023_ICCV}. However, a recurring challenge with LLMs is their tendency to hallucinate, generating plans that are not physically feasible \cite{knowno2023}. To enhance robustness, LLMs require real world grounding, which can be achieved through feedback from the environment \cite{huang2022inner,10161317,Bhat2024GroundingLF}, integration with visual perception systems \cite{Liu2024MOKAOR,Gao2023PhysicallyGV,zhi2024closedloop}, or human-in-the-loop interventions like question-answering \cite{ZhangLu-RSS-21,10161333}. Vision-Language Models (VLMs), trained on vast image-text datasets, excel at visual reasoning tasks \cite{Yang2023TheDO} and have been applied to diverse robotics problems such as encoding 3D semantic memory \cite{conceptgraphs,pmlr-v229-rana23a}, guiding object manipulation based on language instructions \cite{pmlr-v229-stone23a,pmlr-v164-shridhar22a}, and enabling robotic navigation \cite{Hong_2021_CVPR,10160969}. Recent work has focused on training VLMs using multimodal robotic demonstrations, where vision and language are directly mapped to actions \cite{rt12022arxiv,rt22023arxiv,saycan2022arxiv,driess2023palme}. These methods show strong performance in familiar environments but require substantial data and GPU resources for deployment in novel settings \cite{Wake2023GPT4VisionFR}. Consequently, modular systems that integrate planning, grounding, control, and feedback appear more promising for robust robot automation \cite{Liu2024OKRobotWR, Melnik2023UniTeamOV}.

\noindent \textbf{Referring Grasp Synthesis (RGS):} 
Earlier approaches for RGS often used LSTM networks. INGRESS \cite{shridhar2018interactive} employed two LSTMs for grounding, one generating visual object descriptions and another assessing pairwise relations between candidates. \cite{rao2018learning} introduced a learning-based approach incorporating grasp type and dimension selection models for predicting grasp poses from natural language object descriptions. However, these methods struggled with natural language complexities, hindering precise visual grounding. Recent studies show the effectiveness of VLMs in associating language with images \cite{kirillov2023segany,ren2024grounded}. \cite{Mirjalili2023LANgraspUL} used GPT-4 and Owl-VIT \cite{minderer2022simple} to identify objects for grasping from text queries. \cite{10161041} employed CLIP as a vision-text encoder with cross-modal attention for sampling and scoring grasp poses. \cite{lu2023vl} introduced the RoboRefIt corpus to train a transformer-based network to predict 2D object masks from referred text queries. Neural Radiance Fields (NeRFs) can also used for grounding natural language to 3D directly, followed by grasp pose estimation \cite{shen2023F3RM,lerftogo2023}. However, computing NeRFs is time-consuming and thus difficult for real-world deployment. End-to-end RGS directly maps natural language queries to grasp parameters. \cite{Chen2021AJN} trained a ResNet50-LSTM network for merging multi-modal features for GPE. \cite{Jin2024ReasoningGV} fine-tuned a multi-modal VLM for reasoning expression segmentation along with GPE. \cite{Tang2023TaskOrientedGP} used CLIP multi-modal features to train a fusion network with self and cross-attention for task-oriented grasping. \cite{tziafas2023language} released the OCID-VLG dataset for RGS, fine-tuning a CLIP-based model with a transformer decoder for pixel-level object segmentation and GPE. Robust end-to-end RGS requires diverse annotated datasets with images, text queries, and grasp poses, but such datasets are either limited or focus on a small, fixed object set. Recently, \cite{Vuong_2024_CVPR} created a large-scale dataset using foundational models for end-to-end RGS. However, it relies on 2D grasp poses, which are less robust than 6D grasp poses in real-world cluttered scenes. Recent work in GPE has focused on training models with large and diverse pose datasets, such as GraspNet-1Billion \cite{fang2020graspnet}, and learning robust grasp poses for unseen objects. Our two-stage RGS approach uses pre-trained VLMs to generate accurate pixel-level segmentation of referred objects, which can then be used by state-of-the-art GPEs to generate stable 6-DOF grasp poses.

\noindent \textbf{Visual Grounding: } Visual Grounding (VG) in robotics identifies an object or region in 2D/3D space related to a given query, making it critical for connecting natural language to the real world~\cite{zhao20213dvg, he2021transrefer3d, dendorfer21iccv}. This process involves segmenting the referred part and projecting it across camera views to construct a 3D object point cloud. Downstream grasping modules can then use this point cloud to determine grasp poses \cite{Du2020VisionbasedRG}. Our work focuses on 2D Visual Grounding, which is often studied as Referring Image Segmentation (RIS) in computer vision. Traditional RIS models utilize Convolutional Neural Networks or Long Short-Term Memory Networks \cite{hu2016natural, nagaraja2016modeling, yu2018mattnet}. The field has advanced significantly with transformer-based architectures enhancing language grounding in visual contexts \cite{deng2021transvg, liu2021cross, yang2022improving, feng2021encoder}. State-of-the-art RIS models employ large transformer architectures with cross-attention and fine-tune for generating object bounding boxes or pixel-wise segmentation~\cite{UNINEXT,wu2023uniref}. Such models are often compute intensive, requiring multiple A-100 GPUs for finetuning and deployment making it challenging for usage in real-time processing for robotic visual grounding. PolyFormer~\cite{Liu_2023_CVPR} uses a transformer-based architecture with separate visual and textual feature extractors and a multimodal fusion strategy for polygon regression of the segmentation mask in the image. Annotating accurate polygon regression coordinates for segmentation masks by human experts is time consuming, with each sample on average requiring 79s~\cite{Papadopoulos2017ExtremeCF}. Weakly supervised methods alleviate some of the costs associated with segmentation annotations by employing innovative strategies, such as combining positive and negative queries during training or using negative anchor features \cite{liu2023referring,Jin_2023_CVPR}. Although these models report high performance in diverse testing, their application in robotics face challenges due to a lack of robotics related data representation in popular datasets like Flickr30K-entities \cite{7410660}, RefCOCO \cite{Yu2016ModelingCI} and ReferIt~\cite{kazemzadeh-etal-2014-referitgame}. Robotic setups often contain (i) cluttered environments with overlapping objects and occlusions, and (ii) complex referring queries describing object attributes, such as color, shape, or relative position, to uniquely identify the object to grasp in the presence of distractors. For example the query: \textit{``Grab the blue rectangular box on the right side"} can resolve ambiguity if the workspace contains multiple boxes. Recent work highlights the challenges in directly using RIS methods in robotics, where failure to predict accurate masks for smaller scaled objects and in cluttered scenes causes downstream problems in manipulation~\cite{tziafas2023language,Iioka2023MultimodalDS}. Thus recent autonomous robots like MOKA~\cite{Liu2024MOKAOR} and OK-Robot~\cite{Liu2024OKRobotWR} use open-set detectors like GroundedSAM~\cite{ren2024grounded} and OwlVIT~\cite{minderer2022simple} for visual grounding, as they are more robust to language variations and complexities. These models are trained on millions of images and use transformer-based architectures for generating probabilistic bounding box predictions after sampling text queries for a large set of objects. We identify four critical characteristics of an ideal VG model: (i) ability to leverage referring attributes in the input text to distinguish target object among distractors, (ii) robustness to occlusions and partial visibility of the target object, (iii) ease of fine-tuning on custom annotated datasets of RGB-Text-Mask tuples to improve in-domain performance, and (iv) generalizability to open-vocabulary settings with unseen object categories.

\noindent \textbf{Grasp Synthesis: } 
Robotic grasping has been explored using both 4 and 6 Degree of Freedom (DOF) grasp poses. The 4-DOF grasp representation involves 3D positioning and top-hand orientation about the robot gripper axis \cite{Depierre2018JacquardAL,9561398,5980145}. In contrast, the 6-DOF method, which includes three kinematic variables for both position and orientation, provides greater robustness, allowing object manipulation in cluttered environments with an arbitrary direction of grasping \cite{lu2022hybrid,9126187,9560844}. \cite{fang2020graspnet} introduced the GraspNet-1Billion dataset, which has been used to train DNN-based models on RGB-D or point cloud data \cite{gou2021RGB,lu2022hybrid}. Recently, \cite{fang2023anygrasp} achieved a 93.3\% grasping accuracy by training GSNet \cite{gsnet2020ke} on GraspNet-1Billion, utilizing 3D convolutional layers to process point cloud data, followed by stacked MLP layers to predict grasp parameters. We focus on training a VG model to produce segmented object masks, which combine with depth maps to generate object-level point clouds compatible with downstream grasping modules.

\section{Proposed Method: Hierarchical FiLM - ClipSeg (HiFi-CS)}
\label{sec:method}

We study VG in two scenarios: Closed and Open Vocabulary. In \textit{Closed Vocabulary}, models are tested on datasets with pre-known object categories. \textit{Open Vocabulary} evaluations assess methods on unseen environments and objects.

\begin{figure}[!h]
\centering
\includegraphics[width=0.95\columnwidth]{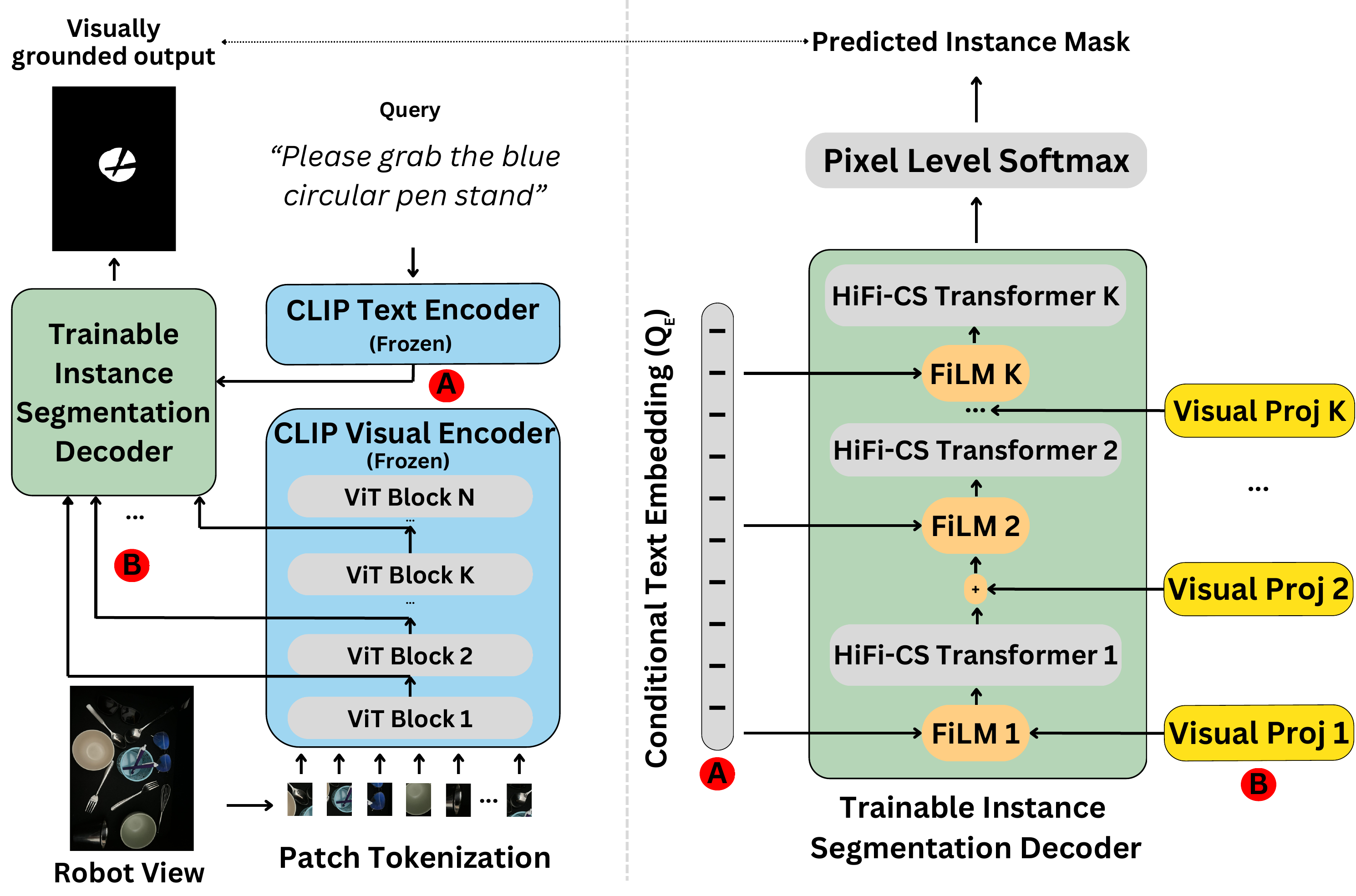}
\caption{HiFi-CS for Robotic Visual Grounding. Left: Blue modules are frozen, we choose K = \{1, 3, 5, 7, 9\}. Right: Zoomed-in view of trainable decoder. ViT: Vision Transformer.}
\label{fig:architecture}
\end{figure}

Referring queries in robotic visual grounding often contain multiple object attributes that need to be accurately remembered and utilized for segmentation mask prediction. An effective visual grounding model must therefore employ a robust multi-modality fusion and learning strategy to correctly identify the target object. Our model (\cref{fig:architecture}) utilizes a frozen CLIP VLM as a feature extractor for both image and text modalities, leveraging its joint embedding space. We hypothesize that a hierarchical and repeated fusion of image and text modalities can provide the segmentation decoder with sufficient clues to learn accurate segmentation masks especially when text queries are longer and more complex. Featurewise Linear Modulation (FiLM) \cite{perez2018film} layers have been shown to effectively merge multi-modal features \cite{Vries2017ModulatingEV}. Building on this concept, we integrate FiLM layers into our trainable segmentation decoder.

In our approach, the referring text query is processed through the CLIP text encoder to produce conditional text embeddings, while the RGB image of the workspace passes through the CLIP visual encoder, with projections extracted from a selected set of K transformer blocks. We introduce FiLM layers to fuse extracted visual projections with conditional text embeddings before each decoder block, enhancing semantic retention for disambiguated segmentation. Our lightweight segmentation decoder, which contains K transformer blocks, receives language-conditioned visual inputs from FiLM. The final output, resized to match the input image dimensions, is then processed through a softmax layer. This layer assigns a binary label to each pixel, predicting whether it belongs to the referred object or the surrounding background.
Our work differs from previous methods which typically merge multi-modal features at the first step of the decoder~\cite{lueddecke22_cvpr}. We show that continually merging these features within the segmentation decoder effectively improves semantic retention without parameter-heavy cross-modal attention. As a result, our model only contains around 6M trainable parameters, a 100x reduction over previous methods in Robotic VG and RIS. 

\noindent \textbf{Mathematical Formulation: } Given input RGB image I $\in$  $\mathbb{R}^{H \times W \times 3}$ and referring query Q, VG models predict a binary mask M $\in$ $[0,1]^{H \times W}$ with H and W denoting the height and width of the image, respectively. The region with pixel values equal to 1 corresponds to the object referred to by Q, while pixel values equal to 0 correspond to the background. CLIP intermediate projections $(P_1, P_2, \dots P_K)$ are combined with query embedding $Q_E$ using FiLM layers to generate decoder inputs $(D_1, D_2, … D_K)$ -
\begin{align}
    D_i = \alpha(Q_E) \cdot (P_i + T_{i-1}(D_{i-1})) + \beta(Q_E)   \ \ 
\end{align}
where $\alpha$ and $\beta$ are feed-forward networks, $T_{i-1}$ denotes the $(i-1)^{th}$ transformer block and $D_0 = 0$. Decoder progressively learn representations to segment the correct object. The final decoder output \( D_K \) is upsampled to the original image resolution \( H \times W \) with the predicted mask given by:
\begin{align}
M_{\text{pred}} = \text{Softmax}(\text{TransConv2D}(D_K)). 
\end{align}

\noindent \textbf{Boosting Zero-Shot Performance: } We freeze the pre-trained CLIP VLM and only train the decoder. CLIP, pre-trained with contrastive image-language learning on large internet scale dataset, generates high quality visual and textual embeddings \cite{Radford2021LearningTV}. By leveraging these pre-trained capabilities, we hypothesize improved performance on unseen objects compared to full fine-tuning-based methods.

\section{Experimental Results}
\label{sec:result}

This section discusses experiments across Closed and Open-Vocabulary settings. 

\noindent \textbf{Evaluation Metric: } We use Intersection Over Union (IoU), averaged over test sets, to evaluate models along with thresholded precision scores. Given a predicted segmentation mask M and ground truth mask G such that M, G $\in$ $[0,1]^{H \times K}$, IoU is calculated as the intersection of M and G divided by their union. P@X scores the percentage of predictions with IoU higher than threshold X. 

\noindent \textbf{Referring Text Complexities:} Referring queries often include various object attributes such as color, shape, relative position, or inter-object relationships to uniquely identify an object. As the number of attributes increases, so does the complexity of the text query. This requires the visual grounding model to account for all attributes in order to accurately identify the object to be retrieved. To quantify the complexity of referring queries, we use Named Entity Recognition (NER), which associates each word in a sentence with predefined entities like object names, colors, shapes, sizes, etc. Specifically, we employ a state-of-the-art NER model, GLiNER \cite{zaratiana2023gliner}, to categorize our test set into four groups based on the number of attributes present in the query. This allows us to compute IOU scores at varying levels of query complexity. For instance, the query, ``Please grab the blue pen on the right side,” contains three attributes (object = pen, color = blue, position = right). More examples are provided in the supplementary material (Section 1).

\subsection{Datasets}
\label{subsec:dataset}
We select two recent VG datasets for closed-vocabulary experiments, featuring cluttered indoor images with graspable objects and multiple instances, suitable for robotics.

\noindent \textbf{RoboRefIt \cite{lu2023vl} }consists of 187 distinct real-world indoor scenes with 66 unique object categories. The resulting corpus contains around 50K tuples (RGB, text, mask). Two test splits are provided: Test A comprises samples with seen object categories as in the training set, while Test B comprises samples with unseen object categories. This corpus does not contain any annotations for grasp parameters, and thus can only be used for training visual grounding models. 

\noindent \textbf{OCID-VLG \cite{tziafas2023language} }comprises 1763 highly cluttered indoor tabletop scenes and 31 unique graspable objects. Many scenes contain multiple instances of the same object and thus text queries use attributes such as object color, shape, relative position, and spatial relationships. The final dataset consists of roughly 89K (RGB, text, mask) tuples. Each tuple is also annotated with grasp parameters, but we only use the visual grounding masks for our experiments.

\subsection{Baselines}

We use two competitive VG baselines and two open-set RIS detectors for thorough comparison, addressing a gap in previous works to identify the best method for robotic VG.

\noindent \textbf{VL-Grasp (VL-Gr): } Introduced in \cite{lu2023vl}, VL-Grasp consists of a BERT text encoder and ResNet50 image encoder. The encoded output is concatenated and passes through a visual-lingual transformer with cross-modal attention. Finally, a decoder predicts a pixel-wise segmentation map. We train VL-Gr separately on RoboRefIt and OCID-VLG.

\noindent \textbf{CROG:} Similar to our method, CROG \cite{tziafas2023language} also uses the CLIP visual and text encoder to generate embeddings for referring text and RGB image. These embeddings pass through a multi-modal feature pyramid network and cross-modal attention layers, leading to a pixel-wise segmentation decoder. In contrast to our method, CROG finetunes the entire network including the CLIP layers. Although this baseline includes a decoder for predicting grasp parameters, we only use the VG part by eliminating the grasp loss. CROG is trained separately on RoboRefIt and OCID-VLG.

\noindent \textbf{GroundedSAM (GrSAM):} This is a zero-shot baseline that combines an open-set object detector (Grounded DINO \cite{liu2023grounding}) and a powerful segmentation model (SAM \cite{kirillov2023segany}). Grounded DINO takes as input the RGB image and a text query, outputting a bounding box over the predicted object. This passes through SAM to generate a segmentation mask. GrSAM demonstrated high performance on open-set object detection, and we use it without any further training \cite{ren2024grounded}.

\noindent \textbf{OwlViT + SAM (OwlSAM):} This is another zero-shot baseline that combines an open-set object detector (OwlVIT \cite{minderer2022simple}) with SAM. Similar to GroundedSAM, we use it without any further training as OwlVIT is trained on large datasets across diverse domains of visual grounding tasks.

\subsection {Experimental Setup}

Our lightweight model is trainable on a single RTX 5000 GPU whereas VL-Gr and CROG require four GPUs in parallel. Training employs pixel-wise binary cross-entropy loss, Adam optimizer~\cite{Kingma2014AdamAM}, and a cosine learning rate scheduler. Zero-shot baselines use GPU-accelerated inference.

\subsection{Closed Vocabulary}

\begin{table}[!h]
\setlength{\tabcolsep}{0.5pt}
\begin{center}
 \begin{tabular}{c c c c c c c c}
 \toprule[1pt]
 \textbf{Model} & \makecell{\textbf{Test}\\\textbf{(Seen)}} & \makecell{\textbf{Test}\\\textbf{(Unseen)}} & \textbf{P@50} & \textbf{P@60} & \textbf{P@70} & \textbf{P@80} & \textbf{P@90}\\ 
\midrule[1pt]
 GrSAM & - & 45.87 & 46.88 & 41.22 & 38.80 & 35.85 & 24.14 \\
 
 OwlSAM & - & 41.39 & 50.58 & 49.93 & 46.71 & 42.93 & \textbf{28.80} \\
 \midrule[0.5pt]
 VL-Gr & 85.46 & 60.89 & 68.21 & 63.38 & 57.33 & 47.86 & 24.55 \\
 
 CROG & 75.46 & 61.84 & 77.89 & 71.79 & 55.55 & 24.02 & 0.60 \\
 \midrule[0.5pt]
 \textbf{HiFi-CS} & \textbf{85.73} & \textbf{70.74} & \textbf{79.74} & \textbf{74.70} & \textbf{66.58} & \textbf{52.31} & 25.41 \\

 \bottomrule[1pt]
 \end{tabular}
\end{center}
\vspace{-10pt}
\caption{Performance of VG models on RoboRefIt. Zero-shot accuracies are listed under test unseen column since all samples are unseen. All scores in IOU. P@X is calculated on test unseen.}
\label{tab:roborefit-results1}
\vspace{-10pt}
\end{table}

Our model outperforms all baselines on the RoboRefIt corpus (\cref{tab:roborefit-results1}). The performance gap of VL-Gr and CROG between seen and unseen objects is substantial (25\% and 14\% respectively), indicating likely over-fitting to seen object categories. Our method achieves improved performance due to a compact and streamlined architecture design that leverages the strengths of frozen multi-modal embeddings from a VLM like CLIP. While CROG also uses CLIP as the backend feature extractor for the image and text modalities, it trains the entire CLIP model and thus loses the benefits of pre-training the VLM on millions of real-world images. CLIP was pre-trained with a contrastive loss to map images to their corresponding descriptive captions and, as a result, learns to transform an image and its corresponding caption to closer locations in the joint embedding space. Text queries in Referring Grasp Synthesis describe the referred object using multiple attributes like object category, color, shape, position, etc. By mapping these queries to the CLIP embedding space, the resulting multi-modal features are rich in semantics about the referred object. Thereafter, a hierarchical application of FiLM to fuse the embeddings and pass through a sufficiently large decoder effectively learns mappings to a pixel-level segmentation mask.

\begin{table}[!h]
\setlength{\tabcolsep}{2pt}
\begin{center}
 \begin{tabular}{c c c c c c c c}
 \toprule[1pt]
 \textbf{Model} & \textbf{Test} & \textbf{P@50} & \textbf{P@60} & \textbf{P@70} & \textbf{P@80} & \textbf{P@90}\\ 
\midrule[1pt]
 GrSAM & 29.39 & 23.45 & 18.23 & 16.95 & 14.55 & 5.69 \\
 
 OwlSAM & 19.92 & 22.21 & 21.56 & 20.28 & 17.18 & 6.35 \\
 \midrule[0.5pt]
 VL-Gr & 87.35 & 94.19 & 91.63 & 86.27 & 63.15 & 50.19 \\
 
 CROG & 78.89 & \textbf{97.09} & \textbf{95.27} & 84.64 & 58.74 & 10.53 \\
 \midrule[0.5pt]
 \textbf{HiFi-CS} & \textbf{88.26} & 92.68 & 92.13 & \textbf{91.53} & \textbf{89.69} & \textbf{83.21} \\

 \bottomrule[1pt]
 \end{tabular}
\end{center}
\vspace{-10pt}
\caption{Performance of VG models on OCID-VLG Corpus. We used a 70-30 train-test split to compute the test IOU scores.}
\label{tab:ocidvlg-results1}
\end{table}

\begin{figure*}[!ht]
    \centering
    \begin{subfigure}[b]{0.49\textwidth}
        \includegraphics[width=\textwidth]{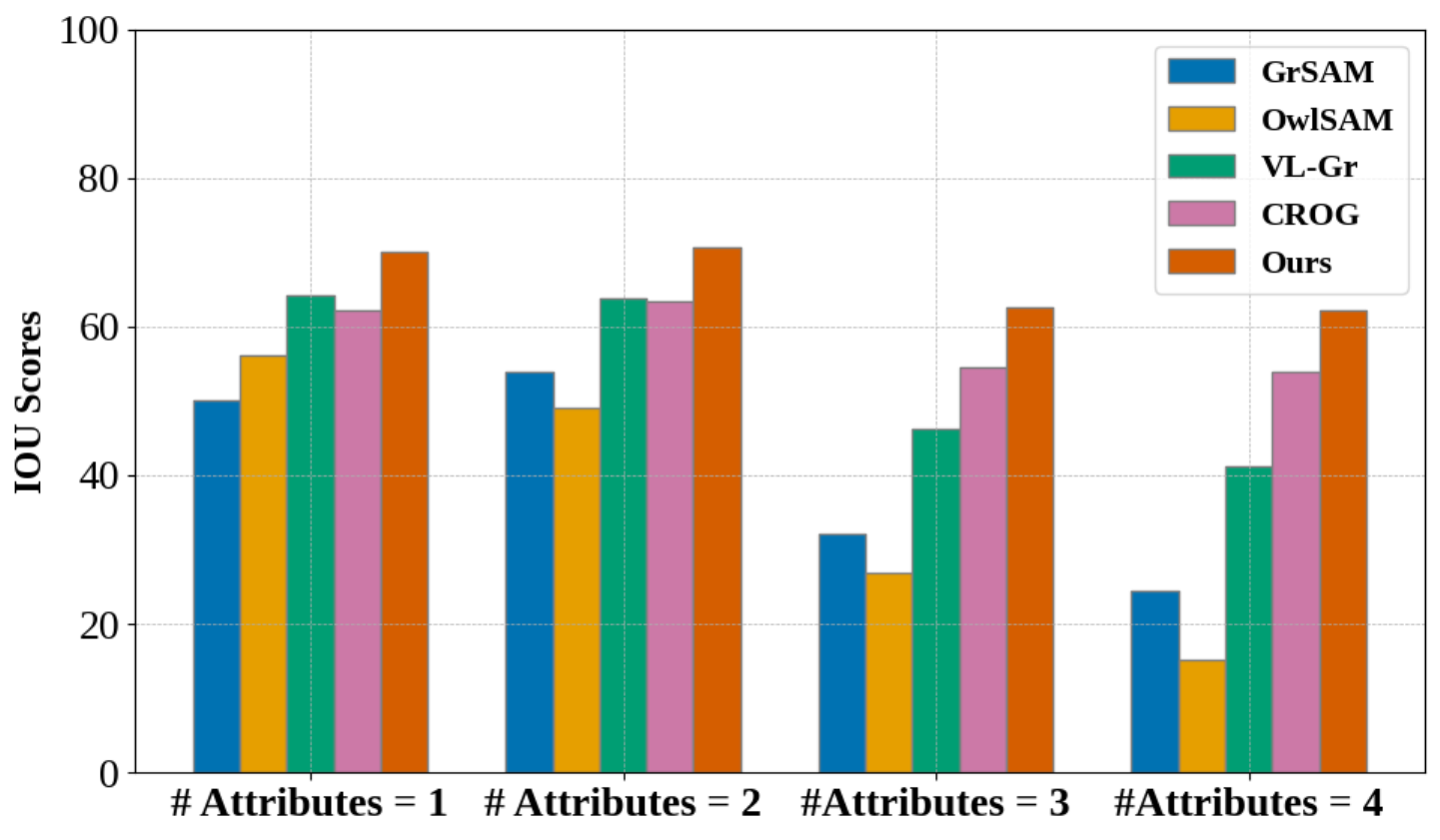}
        \caption{Attribute wise IOU scores on RoboRefIt}
        \label{fig:roborefit-attr-scores}
    \end{subfigure}
    \hfill 
    \begin{subfigure}[b]{0.49\textwidth}
        \includegraphics[width=\textwidth]{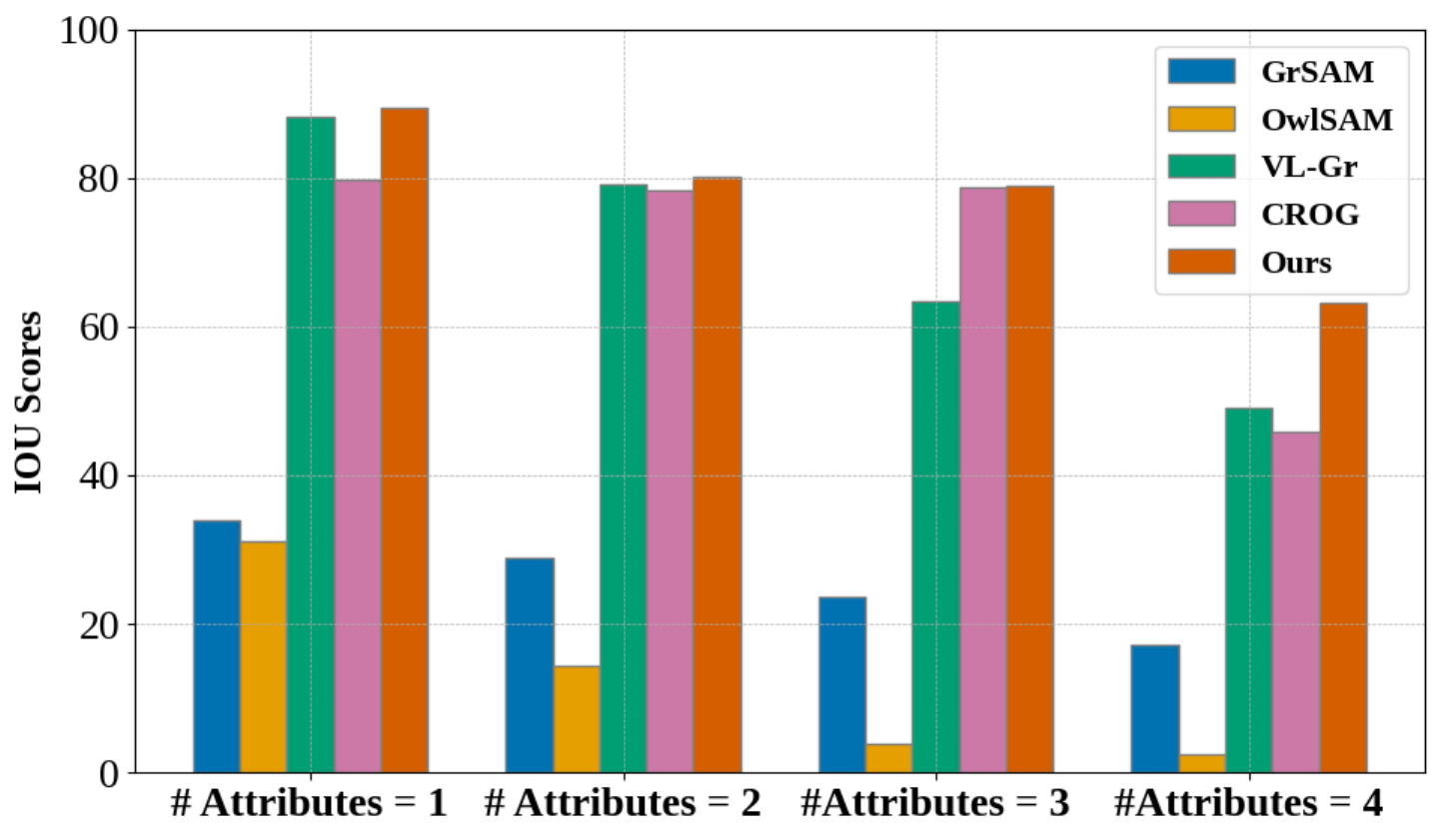}
        \caption{Attribute wise IOU scores on OCID-VLG}
        \label{fig:ocidvlg-attr-scores}
    \end{subfigure}
    \caption{Comparing Visual Grounding baselines across text queries. More attributes increase complexity, requiring the instance mask to be conditioned on properties like color, shape, and position.}
    \label{fig:attr-scores}
\end{figure*}

\begin{table*}[!htpb]
\setlength{\tabcolsep}{3pt}
\renewcommand{\arraystretch}{0.85}
\begin{center}
 \begin{tabular}{c c c c c c c c c c c c c}
 \toprule[1pt]
 \textbf{Model} &  \textbf{Size} & \textbf{Inference Time} & \textbf{IOU} & \textbf{A=1} & \textbf{A=2}  & \textbf{A=3} & \textbf{A=4} & \textbf{P@50} & \textbf{P@60} & \textbf{P@70} & \textbf{P@80} & \textbf{P@90}\\ 
\midrule[1pt]
GrSAM & 172M+308M & 0.44s & 41.65 & 44.67 & 43.71 & 34.64 & 14.99 & 52.17 & 50.52 & 48.44 & 46.71 & 43.85 \\
 OwlSAM & 88M+308M & 0.40s & 41.88 & 47.34 & 39.71 & 39.78 & 24.94 & 42.60	& 41.25	& 39.41	& 37.95	& 35.2\\
 \midrule[0.5pt]
 VL-Gr & 88M & 0.42s & 15.24 & 9.37 & 17.26 & 18.61 & 41.36 & 17.65	& 15.19	& 11.71	& 8.50	& 1.88\\
 CROG & 150M & 0.82s & 16.89 & 8.93 & 17.68 & 27.05 & 31.65 & 18.18	& 16.80	& 12.93	& 9.77	& 2.26\\
 HiFi-CS & \textbf{6M} & \textbf{0.32s} & 22.56 & 15.62 & 22.20 & 33.23 & 41.38 & 23.92	& 19.45	& 15.15	& 10.76	& 3.1\\
 \midrule[0.5pt]
 \textbf{\makecell{GrSAM\\+HiFi-CS}} & \makecell{172M+308M\\+6M} & 0.46s & \textbf{52.77} & \textbf{51.41} & \textbf{51.21} & \textbf{62.65} & \textbf{45.12} & \textbf{54.16}	& \textbf{52.60}	& \textbf{50.35}	& \textbf{48.96}	& \textbf{45.06}\\
 \bottomrule[1pt]
 \end{tabular}
\caption{Zero shot evaluation on RoboRES. A=n denotes the subset of test set with n referring attributes in each query.}
\label{tab:robores-zs-results}
\end{center}
\vspace{-20pt}
\end{table*}

Similar results are obtained in the OCID-VLG corpus (\cref{tab:ocidvlg-results1}) where our model improves significantly at higher precision threshold, demonstrating precise visual grounding and high in-domain performance after training. We observed that larger open-set detectors GroundedSAM and OwlSAM under-perform trained models, highlighting the challenges in directly using these methods in robotic VG. These models suffer when text query complexity increases, whereas HiFi-CS utilizes referring attributes to accurately identify the target object. \cref{fig:attr-scores} shows the performance of all models at increasing levels of text query complexities. All trained models surpass open-set detectors in closed vocabulary settings. For a fixed set of objects, RGS would benefit from trained VG approaches, encouraging data creation for superior in-domain performance.  Ablation studies are provided in the supplementary material (Section 2).

\subsection{Open Vocabulary}
\label{subsec:open-vocab-exp}

Robots must grasp unseen objects in the real world, posing challenges due to the infinite variety of shapes and sizes of graspable objects. We address this by comparing models trained on RoboRefIt with open-set detection models in a zero-shot setting using a new, challenging corpus.

\noindent \textbf{Data Creation: } Given an RGB image, SAM \cite{kirillov2023segany} can segment all objects in the image. However, not all mask outputs correspond to meaningful objects. We collect a corpus of 120 cluttered environment images, manually validate segmentations produced by SAM, and crowd-source the (RGB-Mask) pairs to annotate referring text. Resulting corpus is called \textit{RoboRES} (See supplementary material - Section 3).

\noindent \textbf{Improving open-set detection with language-conditioned guidance: }We introduce a new method for zero-shot inference that leverages the capabilities of both language-conditioned segmentation and open-set detection models. During runtime, prediction from HiFi-CS is compared with the top three predictions of an open-set detection model. The entity with maximum overlap with our prediction is chosen as the output. We choose GroundedSAM as the open-set model and call this approach: GroundedSAM + HiFi-CS (GrSAM+HiFi-CS). 

\begin{table*}[!ht]
    \setlength{\tabcolsep}{5pt}
    \renewcommand{\arraystretch}{0.85} 
    \begin{center}
    \begin{tabular}{c c c c c c c c c c c c c c}
        \toprule[1pt]
        \multirow{3}{*}{\textbf{Model}} & \multirow{2}{*}{\textbf{Level}} & \multicolumn{2}{c}{\textbf{Fruit}} & \multicolumn{2}{c}{\textbf{Soda}} & \multicolumn{2}{c}{\textbf{Container}} & \multicolumn{2}{c}{\textbf{Spray }} & \multicolumn{2}{c}{\textbf{Hardware}} & \multirow{3}{*}{\textbf{Ov-SA}} & \multirow{3}{*}{\textbf{Ov-GA}}\\
        \cmidrule(lr){3-12}
         &  & SA  & GA & SA & GA & SA & GA & SA & GA & SA & GA &  & \\
        \midrule[1pt]

        \multirow{4}{*}{GrSAM} & 1 & 100 & 40 & 80 & 60 & 50 & 40 & 100 & 40 & 70 & 60 & \multirow{4}{*}{75.33} & \multirow{4}{*}{44.00} \\
        \cmidrule(lr){2-12}
        & 2 & 100 & 60 & 100 & 60 & 50 & 0 & 25 & 20 & 65 & 20 &  & \\
        \cmidrule(lr){2-12}
        & 3 & 95 & 60 & 100 & 20 & 50 & 60 & 65 & 40 & 80 & 60 &  & \\
        \cmidrule(lr){2-12}
         & Ov & 98.33 & 53.33 & 93.33 & \textbf{53.33} & 50 & 33.33 & 63.33 & 33.33 & 71.67 & 46.67 &  & \\
         \midrule[1pt]

        \multirow{4}{*}{HiFi-CS} & 1 & 100 & 60 & 75 & 40 & 75 & 40 & 100 & 80 & 80 & 60 & \multirow{4}{*}{85.00} & \multirow{4}{*}{42.67} \\
        \cmidrule(lr){2-12}
        & 2 & 100 & 60 & 100 & 20 & 100 & 20 & 130 & 0 & 100 & 80 &  & \\
        \cmidrule(lr){2-12}
        & 3 & 100 & 40 & 100 & 0 & 75 & 20 & 40 & 40 & 100 & 80 &  & \\
        \cmidrule(lr){2-12}
         & Ov & \textbf{100.00} & 53.33 & 91.66 & 20.00 & \textbf{83.33} & 26.67 & 56.67 & 40.00 & \textbf{93.33} & 73.33 &  & \\
         
         \midrule[1pt]

         \multirow{4}{*}{\textbf{\makecell{GrSAM\\+HiFi-CS}}} & 1 & 100 & 80 & 100 & 60 & 65 & 40 & 85 & 40 & 75 & 80 & \textbf{\multirow{4}{*}{90.33}} & \textbf{\multirow{4}{*}{60.33}} \\
        \cmidrule(lr){2-12}
        & 2 & 100 & 60 & 100 & 20 & 85 & 40 & 100 & 60 & 100 & 100 &  & \\
        \cmidrule(lr){2-12}
        & 3 & 95 & 80 & 100 & 40 & 75 & 60 & 75 & 80 & 100 & 80 &  & \\
        \cmidrule(lr){2-12}
         & Ov & 98.33 & \textbf{73.33} & \textbf{100} & 40 & 75 & \textbf{47} & \textbf{86.67} & \textbf{60} & 91.66 & \textbf{86.67} &  & \\
         \bottomrule[1pt]
         
    \end{tabular}
    \end{center}
    \vspace{-10pt}
    \caption{Results from real-world experiments: Ov (Overall accuracy), SA (Segmentation Accuracy), GA (Grasping Accuracy), all reported as percentages. Scores for object categories and the model are averaged across views, difficulty levels, and categories.}
    \label{tab:real-world-exp}
    \vspace{-10pt}
\end{table*}

\noindent \textbf{Findings: } \cref{tab:robores-zs-results} presents results of testing all models on RoboRES. HiFi-CS outperforms fine-tuned baselines, demonstrating improvements in zero-shot performance. As a smaller model, HiFi-CS averages 0.32 seconds per sample, making it the fastest baseline. It also shows strong performance at higher complexity levels (A=4). However, open-set detectors outperform fine-tuned language-conditioned segmentation models. This is expected, as models like GrSAM and OwlSAM are pre-trained on additional datasets for general segmentation tasks and likely encountered objects similar to our test set. As text complexity increases, performance of open-set detection models declines, while HiFi-CS continues to improve. A hybrid approach, combining GrSAM with HiFi-CS, capitalizes on the strengths of both techniques, resulting in significant improvements. Due to lightweight size of HiFi-CS, inference remains efficient when integrated with GrSAM.

\section{Real World Experiments}
\label{sec:real-world}

We implemented a pipeline of visual grounding and grasping for our experiments. Visual grounding converted natural language instructions to object masks in RGB-D. The projected object level depth maps were used by the pre-trained AnyGrasp SDK \cite{fang2023anygrasp} for generating candidate grasps. We use three VG baselines from our previous experiments for comparing performance in a real robot setting - HiFi-CS, GroundedSAM and GroundedSAM+HiFi-CS.

\begin{figure}[!ht]
  \centering
  
\includegraphics[width=0.45\textwidth]{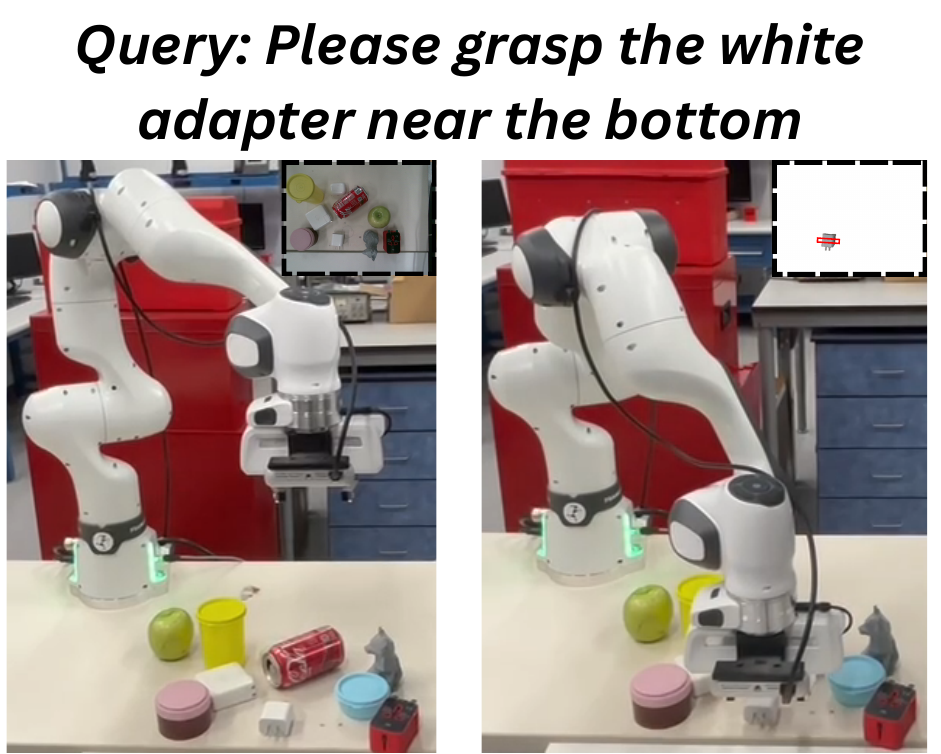}
  \caption{Language Guided Object Manipulation. Left: Robot captures top view image. Right: Referred object grasp is executed.}
  \label{fig:real-world-ex}
  \vspace{-15pt}
\end{figure}

\noindent \textbf{Experimental Setup: } We used five object classes: \textit{Fruit, Soda Can, Food Container, Spray Bottle, and Hardware}. The first two categories are seen whereas the latter three are unseen by our pipeline. Object arrangement involved three levels with increasing number of distractors: Level 1 has one instance per object category, Level 2 has two instances per category, and Level 3 has three instances per category. We evaluated our pipeline with natural language commands using physical attributes visible to human eye. Our VG module captures images of the workspace across 5 views and predicts object masks. Top view mask was provided to AnyGrasp to output grasp poses. Experiments used a 7 DOF Franka Research 3 Arm, with RealSense D455 camera mounted at end-effector to capture RGB-D images. Motion and grasp poses were executed using velocity controller. All scenes used a standard table-top setup (\cref{fig:real-world-ex}).

\begin{table*}[!ht]
\renewcommand{\arraystretch}{0.85}
\setlength{\tabcolsep}{1pt}
\centering
\begin{tabular}{|>{\centering\arraybackslash}m{3cm}|>{\centering\arraybackslash}m{3.5cm}|>{\centering\arraybackslash}m{3.5cm}|>{\centering\arraybackslash}m{3.5cm}||>{\centering\arraybackslash}m{3.5cm}|}
\hline
\multirow{2}{*}{\textbf{}} & \multicolumn{4}{c|}{\textbf{Referring Text Queries}} \\ \cline{2-5}
 & Grasp the \textcolor{red}{hardware adapter} & Where is the \textcolor{red}{longer} \textcolor{red}{blue} \textcolor{red}{food container}? &  Please pass me the \textcolor{red}{coke soda can} & Can you grab the \textcolor{red}{red} \textcolor{red}{apple} on the \textcolor{red}{right}?\\ \hline
 \textbf{Original RGB Image} & \includegraphics[width=3cm]{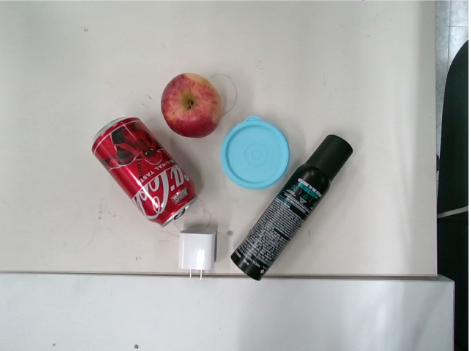} & \includegraphics[width=3cm]{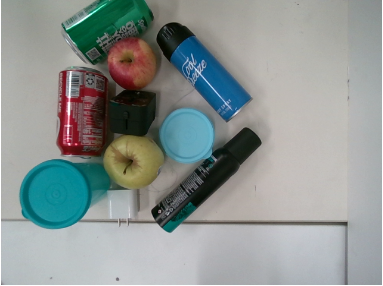} & \includegraphics[width=3cm]{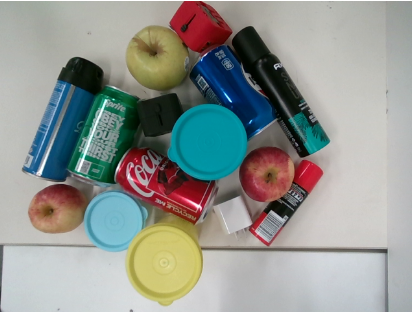} & \includegraphics[width=3cm]{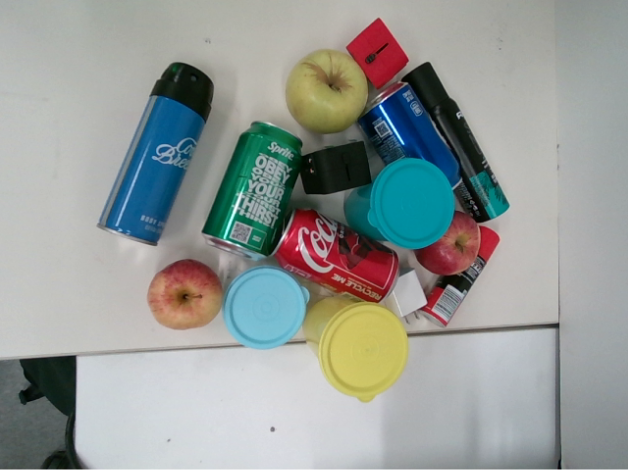} \\ \hline
  \textbf{Ground Truth Mask} & \includegraphics[width=3cm]{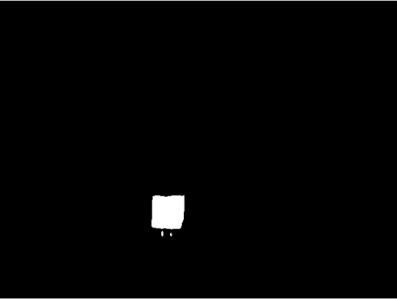} & \includegraphics[width=3cm]{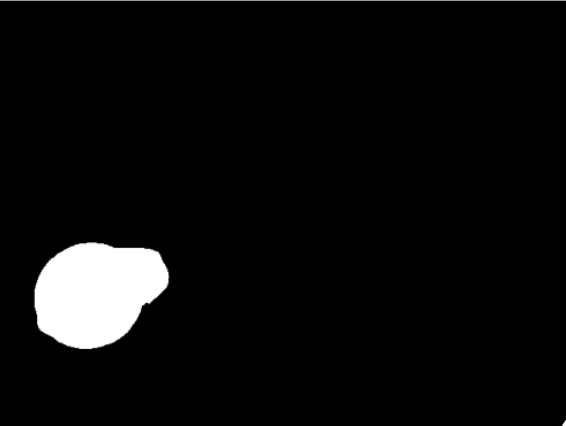} & \includegraphics[width=3cm]{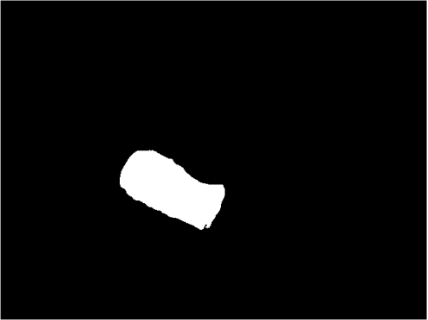} & \includegraphics[width=3cm]{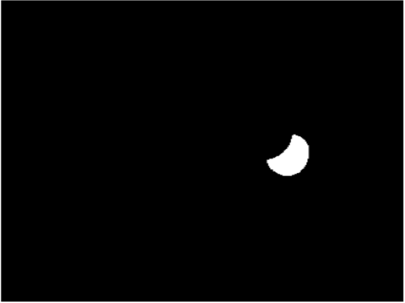} \\ \hline
\textbf{HiFi-CS} & \includegraphics[width=3cm]{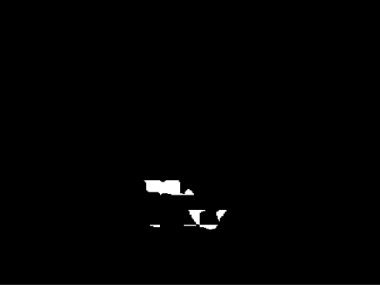} & \includegraphics[width=3cm]{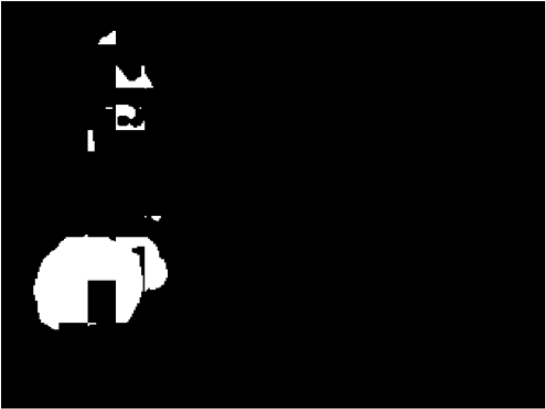} & \includegraphics[width=3cm]{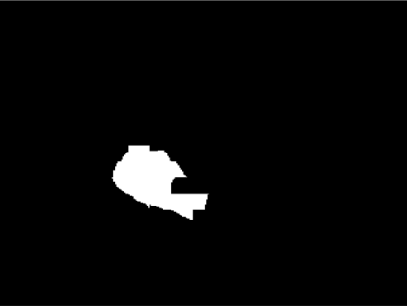} & \includegraphics[width=3cm]{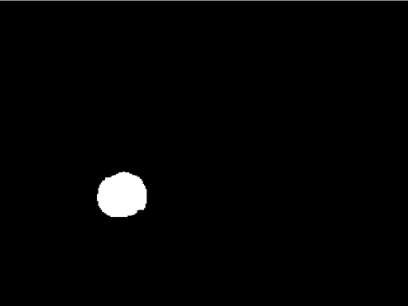} \\ \hline
\textbf{GrSAM} & \includegraphics[width=3cm]{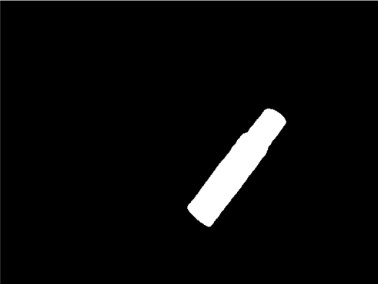} & \includegraphics[width=3cm]{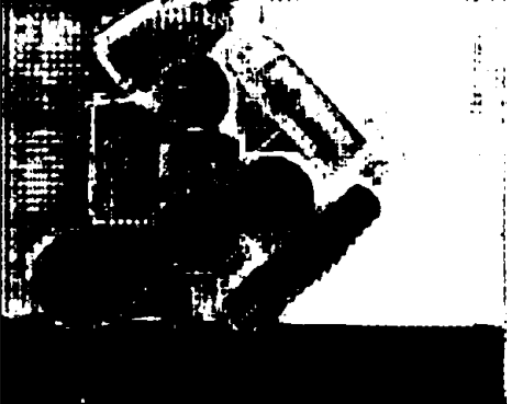} & \includegraphics[width=3cm]{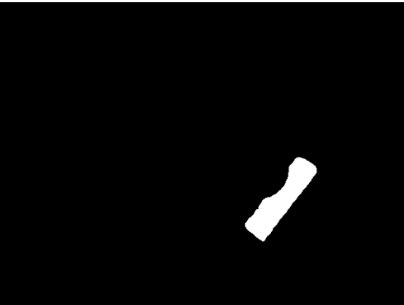} & \includegraphics[width=3cm]{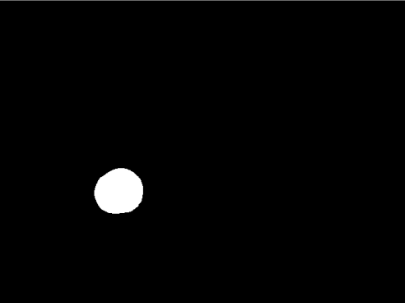} \\ \hline
\textbf{GrSAM+HiFi-CS} & \includegraphics[width=3cm]{figures/grsamhifics-sample1.png} & \includegraphics[width=3cm]{figures/grsamhifics-sample2.png} & \includegraphics[width=3cm]{figures/grsamhifics-sample3.png} & \includegraphics[width=3cm]{figures/grsam-sample4.png} \\ \hline

\end{tabular}
\caption{Qualitative analysis of VG outputs in real-world experiments. The first three examples show improved predictions using GrSAM+HiFi-CS. The last example highlights challenges in grounding complex referring queries, with attributes marked in red.}
\label{tab:real-world-qual}
\vspace{-10pt}
\end{table*}

\noindent \textbf{Findings: } We used two metrics for evaluation: Segmentation Accuracy (SA) and Grasping Accuracy (GA), both scored through visual inspection. SA is 100 if a minimal referring query correctly segmented the required object, and we apply a penalty of 25 each time an additional attribute is required. SA is 0 if the model was unable to identify the object to grab. GA is 100 if the final grasp poses results in successful grasping and lifting of the object, otherwise, GA is 0. Grasping accuracy depends on the segmentation model, as an accurate segmentation mask increases the likelihood of a successful grasp. \cref{tab:real-world-exp} shows the results obtained. Our proposed open vocabulary solution, which combines GroundedSAM with HiFi-CS, outperforms all baselines in both Segmentation Accuracy (SA) and Grounding Accuracy (GA). All methods perform worse on unseen objects (Food Container, Spray Bottle, Hardware) compared to seen objects (Fruit, Soda Can). In some trials with unseen objects, HiFi-CS fails to identify the correct object, causing GroundedSAM to default to the larger or more common object, regardless of the referring attributes in the text. RGB images from different views affect grounding performance, with some views requiring additional attributes for disambiguation. Grasping fails when the predicted pose is slightly offset from the object. Visual servoing-based feedback can help reduce these errors \cite{RIBEIRO2021103757}. Since grasping is not our focus, we leave it to future work. Supplementary material (Section 4) contains more details about our setup and analysis. We provide some qualitative comparisons of prediction outputs for all three baselines in \cref{tab:real-world-qual}. When referring queries contain multiple attributes (highlighted in red), open-set detectors fail to accurately identify the object. Using HiFi-CS as a guide improves prediction quality.

\section{Conclusion and Future Work}
\label{sec:conclusion}

This paper provides extensive comparisons of popular visual grounding techniques in closed and open-vocabulary robotic grasping. We introduce a language-conditioned segmentation model to generate object masks from complex text queries. Referred text in robotics often contains multiple object attributes required for accurate segmentation especially in presence of distractors of the target. Our proposed model uses an intuitive multi-modality fusion design to effectively utilize these attributes. Predicted masks can be used to construct object point clouds for grasp pose estimation. Our model outperforms competitive baselines in closed-vocabulary settings and can be combined with an open-set object detection model for open-vocabulary settings. We demonstrate this on a real robot across three difficulty levels. Our results show that language-conditioned models excel with longer text queries and, when paired with open-set detectors, improve zero-shot performance in visual grounding. Future work will focus on merging planning algorithms for open-vocabulary 6 DOF manipulations and adapting our method for visual grounding in navigation.

 \noindent \textbf{Limitations: }Multi-stage RGS is prone to errors, especially when VG misidentifies the target, resulting in incorrect grasps. To mitigate this, we use a hybrid language-conditioned and open-set segmentation model. Additionally, our system relies solely on a hand-mounted camera, and adding base cameras could improve grasp accuracy.


{\small
\bibliographystyle{ieee_fullname}
\bibliography{egbib}
}

\end{document}


\title{HiFi-CS: Towards Open Vocabulary Visual Grounding For Robotic Grasping Using Vision-Language Models - Supplementary Material}

\author{Vineet Bhat, Prashanth Krishnamurthy, Ramesh Karri, Farshad Khorrami \\
New York University\\
Brooklyn, NY, USA\\
{\tt\small vrb9107@nyu.edu}
}
\maketitle

\section{Analyzing Referring Text Attributes With Named Entity Recognition}
\label{app:ner-attr}

\begin{figure}[!h]
  \centering
  \includegraphics[width=0.48\textwidth]{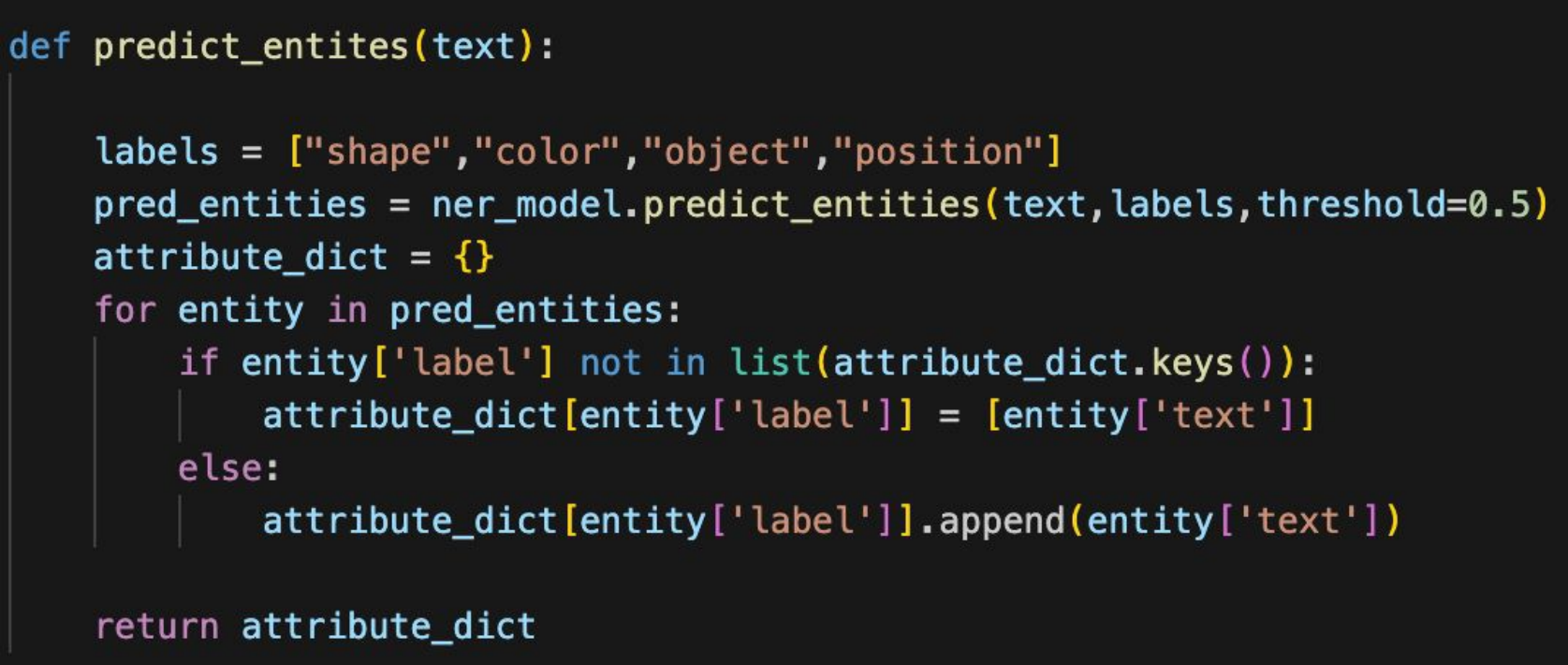}
  \caption{Python function for attribute extraction using GLiNER model (referred as \textit{ner\_model} )}
  \label{fig:ner-pseudocode}
\end{figure}

Named Entity Recognition (NER) is a Natural Language Processing technique used to identify various entities within text input, which is closely related to our attribute analysis. NER models, typically trained on large annotated text datasets, learn to associate each word in a sentence with an entity label such as names, organizations, dates, colors, etc. For identifying and categorizing referring text in our test samples, we utilize the state-of-the-art GLiNER model. Given the referring text, we extract labels for ``Object", ``Color'', ``Shape," and ``Position" using the function illustrated in \cref{fig:ner-pseudocode}. Consequently, test sets are divided into four categories based on the number of attributes extracted by NER (examples provided in \cref{app-tab:attribute_categories}), and metrics are reported separately for each case as detailed in Section 4 of the main paper.

\begin{table*}[h!]
    \centering
    \begin{tabular}{>{\centering\arraybackslash}m{2.2cm} >{\centering\arraybackslash}m{3cm} >{\centering\arraybackslash}m{3.5cm} >{\centering\arraybackslash}m{4cm}}
        \toprule[1pt]
        \textbf{Category} & \textbf{Referring Text} & \textbf{Attributes} & \textbf{Ground Truth} \\
        \midrule[0.5pt]
        \# Attributes = 1 & Where is the multimeter? & \parbox{3.5cm}{1. Object: Multimeter} & \includegraphics[width=0.2\textwidth]{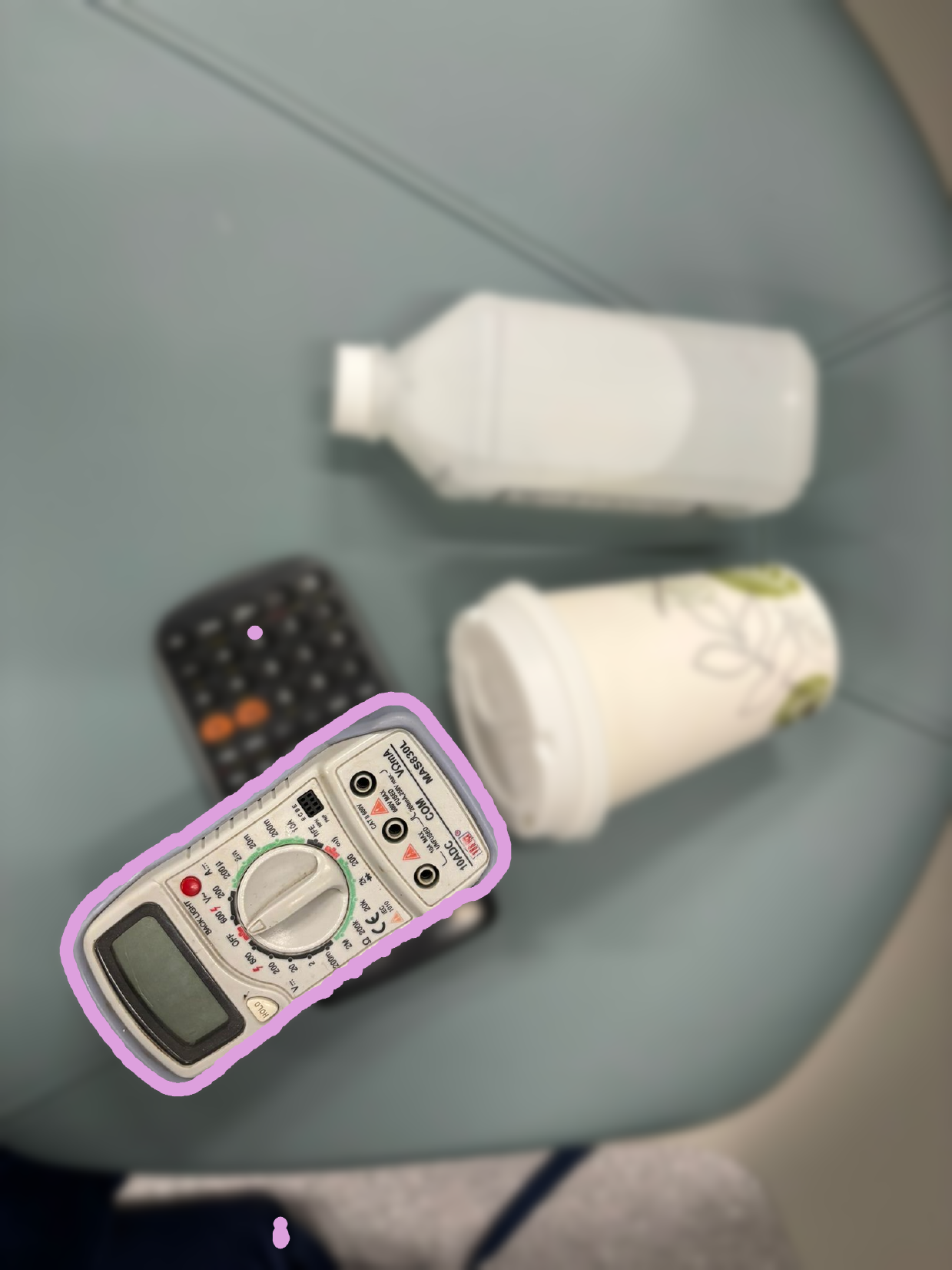} \\
        
        \# Attributes = 2 & Find the coffee cream near the center of the image & \parbox{3.5cm}{1. Object: Coffee cream \\ 2. Position: Center} & \includegraphics[width=0.2\textwidth]{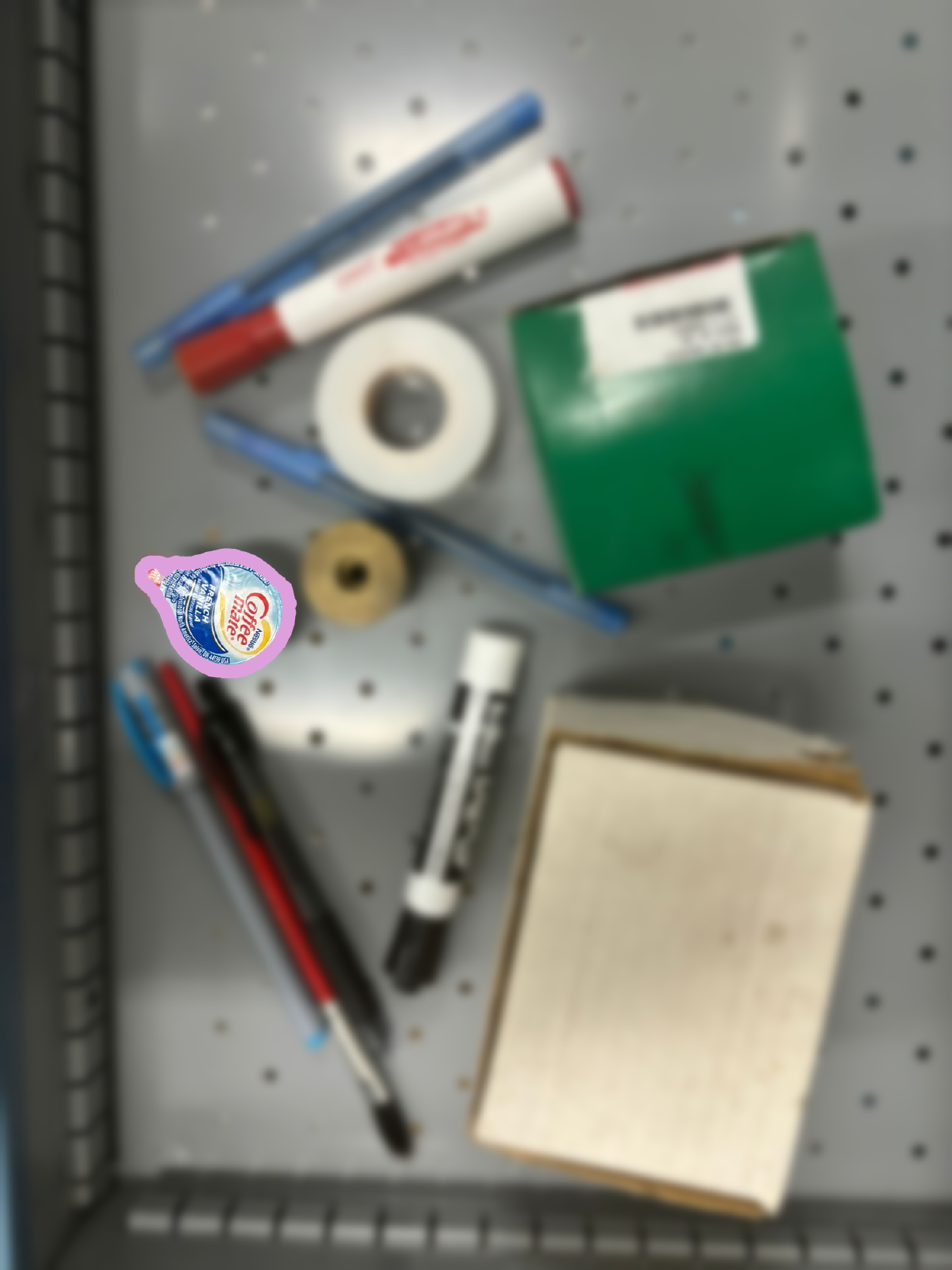} \\
        
        \# Attributes = 3 & Please grab the silver bowl on the left & \parbox{3.5cm}{1. Object: Bowl \\ 2. Color: Silver \\ 3. Position: Left} & \includegraphics[width=0.2\textwidth]{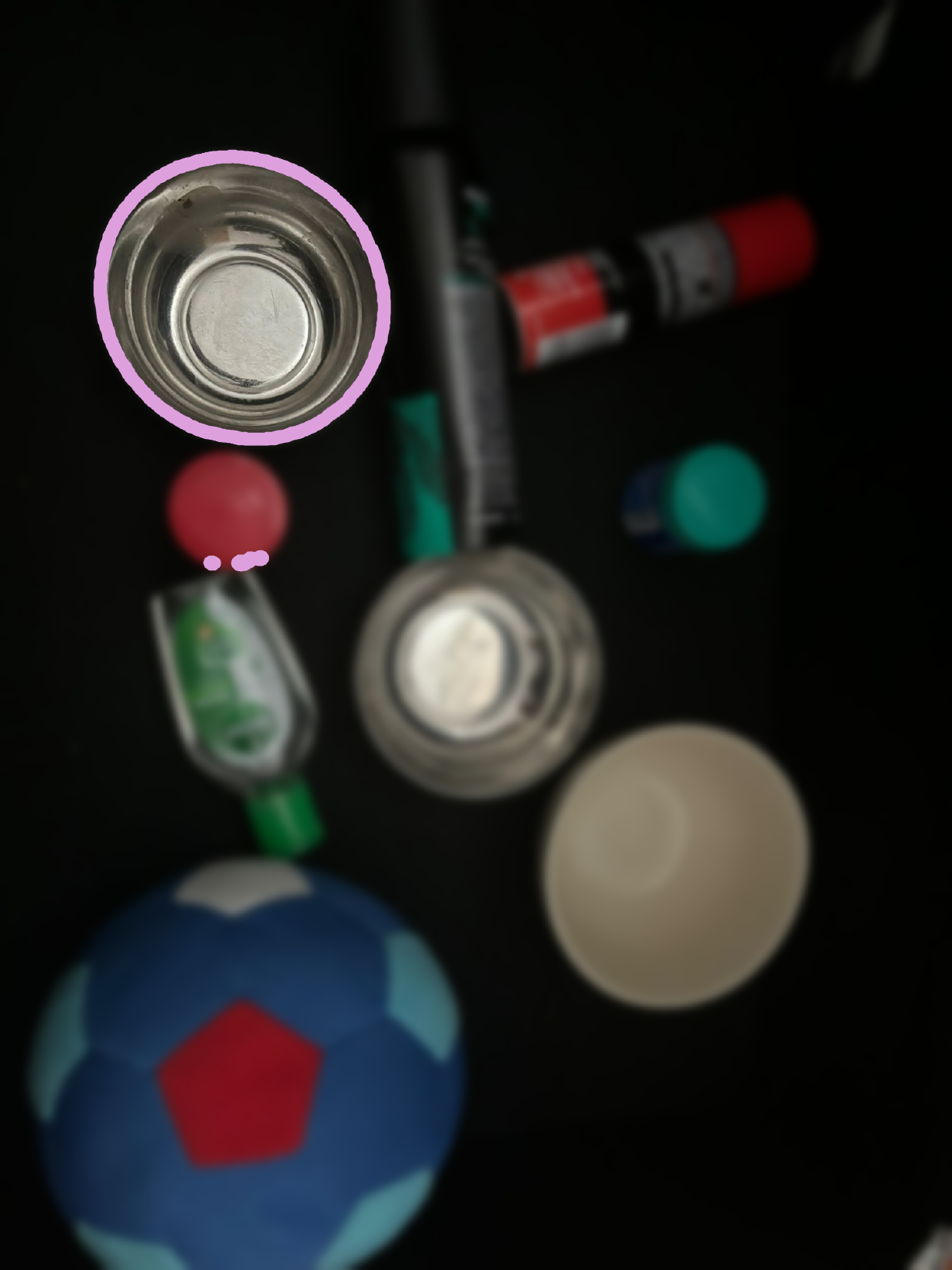} \\
        
        \# Attributes = 4 & Please grab the white circular tape placed on top of the ruler & \parbox{3.5cm}{1. Object: Tape \\ 2. Shape: Circular \\ 3. Color: White \\ 4. Position: Top of ruler} & \includegraphics[width=0.2\textwidth]{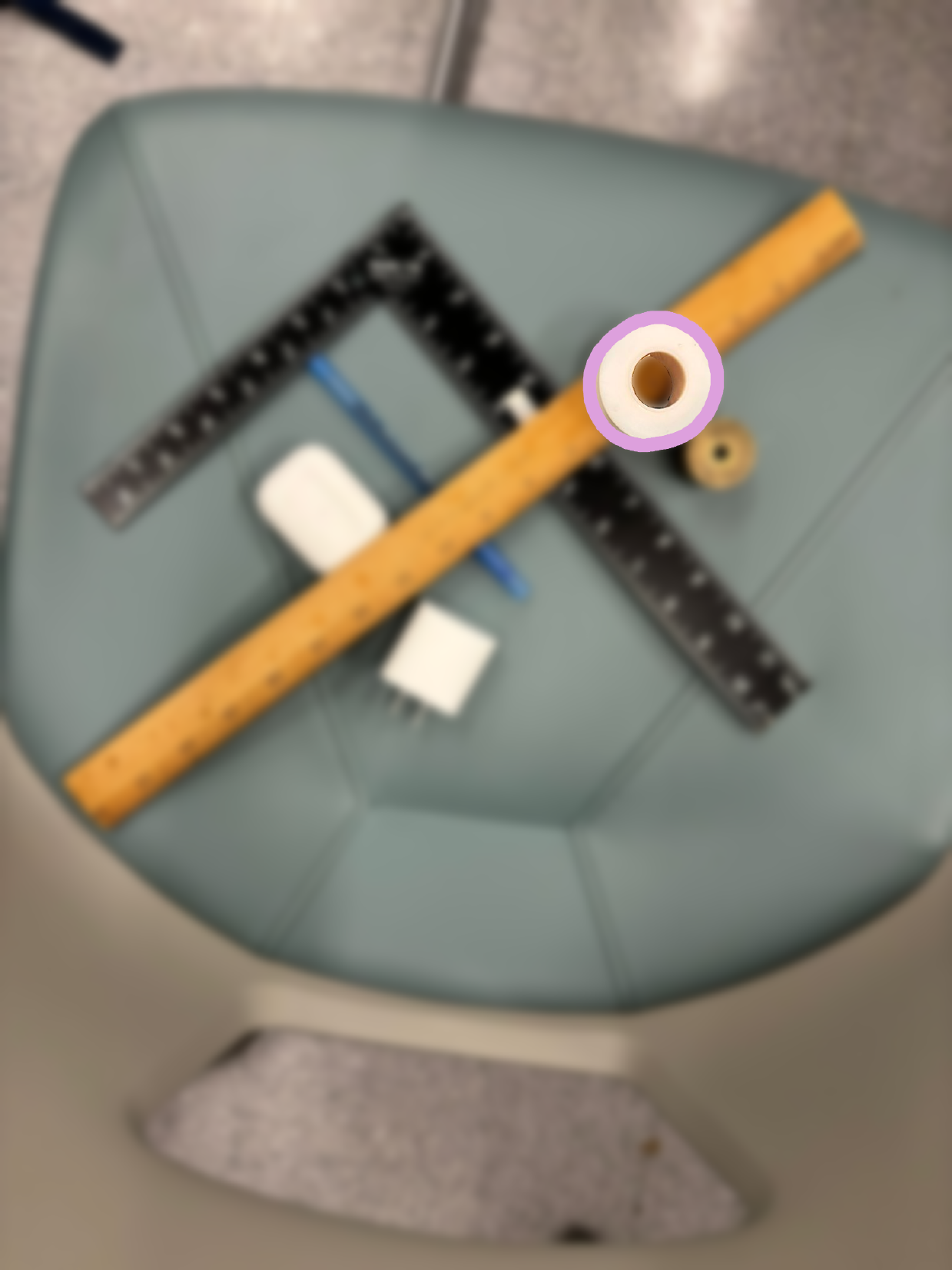} \\
        \bottomrule[1pt]
    \end{tabular}
    \caption{Types of referring text categories covered in this paper. Cluttered scenes demand additional attributes in referring text to uniquely identify objects of interest. Images from the RoboRES corpus.}
    \label{app-tab:attribute_categories}
\end{table*}

\section{Ablation Studies}
\label{app:ablation}

HiFi-CS has various hyperparameters that affect its overall performance. We perform ablation studies on five important aspects of our model: visual projections from the frozen VLM, the dimension of trainable decoder blocks, vision backend used for feature extraction, different types of multimodal fusion strategy and variations in text encoder. All ablations are conducted using the RoboRefIt corpus for consistency, where we report the IoU across seen and unseen objects. 

\begin{table}[!h]
\setlength{\tabcolsep}{3pt}
\centering

 \begin{tabular}{c c c}
 \toprule[1pt]
\textbf{Projection layers} & \textbf{Test Seen IOU} & \textbf{Test Unseen IOU}\\ 
\midrule[1pt]
 K = \{1, 4, 7, 10 \} & 79.45 & 61.12 \\
  \midrule[0.5pt]
 K = \{1, 3, 5, 7, 9\} & 82.45 & 69.61 \\
 \midrule[0.5pt]
 K = \{1, 3, 5, 7, 9, 11\} & 85.41 & 69.37 \\
 \bottomrule[1pt]
 \end{tabular}
\caption{Ablation over number of visual projections extracted. Model iterations below K = 3 gave a poor scores.}

\label{tab:app-ablation-1}
\end{table}

\noindent \textbf{Visual Projections: } The CLIP VLM consists of multiple transformer blocks stacked sequentially. Input patches of the image pass through each transformer block, with each layer learning different levels of semantic information as the input propagates through the model. For a fixed version of CLIP (ViT-B/16), there are 12 transformer blocks in the vision encoder from which we can extract projections. We vary K, the set of transformer blocks chosen, between 4 to 6 to understand the impact on overall performance. This hyperparameter is crucial as the trainable decoder consists of K transformer blocks corresponding to the visual projections. Our results indicate that increasing the number of visual projections enhances performance on seen objects but saturates after K = 5  (Table \ref{tab:app-ablation-1}). Since our objective is to perform well on unseen objects without over-fitting to any test set, we choose K = 5 for our model.

\noindent \textbf{Decoder Dimension: } After FiLM conditioning, the merged multi-modal features pass through decoder transformer blocks. Each block is associated with an embedding dimensionality, which specifies the granularity of intermediate representations to be learned. Increasing the dimensionality also increases the model size. We vary this dimension D between \{64, 128\}. Table \ref{tab:app-ablation-2} presents the results. Increasing the decoder size improves overall performance by allowing the decoder to learn better intermediate representations. However, increasing the size beyond D=128 causes training to diverge, indicating a saturation point.

\begin{table}[!h]
\setlength{\tabcolsep}{3pt}
\centering
 \begin{tabular}{c c c}
 \toprule[1pt]
 \textbf{Decoder Size} & \textbf{Test Seen IOU} & \textbf{Test Unseen IOU}\\ 
\midrule[1pt]
 D = 64 & 82.45 & 66.15 \\
  \midrule[0.5pt]
 D = 128 & 84.80 & 69.59 \\
 \bottomrule[1pt]
 \end{tabular}
\caption{Ablation over dimensionality of the trainable decoder. D denotes to decoder embedding dimension.}

\label{tab:app-ablation-2}
\end{table}

\noindent \textbf{Backend CLIP Vision Transformer: } The official implementation of CLIP provides various vision transformer backends. Larger models typically perform better than base models. We chose two different backend models for our ablation, namely ViT-B/16, and ViT-L/14, where 16 and 14 denote the patch dimensions used in the vision encoder. Table \ref{tab:app-ablation-3} shows our results. As expected, the larger vision transformer backend yields better results across both test splits.

\begin{table}[!h]
\setlength{\tabcolsep}{3pt}
\centering
 \begin{tabular}{c c c}
 \toprule[1pt]
 \textbf{CLIP Backend} & \textbf{Test Seen IOU} & \textbf{Test Unseen IOU}\\ 
\midrule[1pt]
 ViT-B/16 & 84.80 & 69.59 \\
  \midrule[0.5pt]
 ViT-L/14 & 85.73 & 70.74 \\
 \bottomrule[1pt]
 \end{tabular}
\caption{Ablation on vision transformer backends. ViT-L/14 showcases better performance, indicating that our method could improve with larger versions of CLIP backends.}
\label{tab:app-ablation-3}
\end{table}

\noindent \textbf{Multi-Modal Fusion Strategy: } Table \ref{tab:app-ablation-4} shows the results of using two popular fusion methods. While using cross-attention mechanism, we observed that the model reached early saturation, with loss not decreasing even while gradually reducing the learning rate. The computationally expensive cross-attention mechanism might not effectively combine features that lie close in the joint dimension space, whereas a simple FiLM layer maintains the rich semantic information for visual grounding.

\begin{table}[!h]
\setlength{\tabcolsep}{3pt}
\centering
 \begin{tabular}{c c c}
 \toprule[1pt]
 \textbf{MM Fusion Strategy} & \textbf{Test Seen IOU} & \textbf{Test Unseen IOU}\\ 
\midrule[1pt]
 Cross Attention & 83.97 & 63.56 \\
  \midrule[0.5pt]
 FiLM & 85.73 & 70.74 \\
 \bottomrule[1pt]
 \end{tabular}
\caption{Ablation on different Multi-Modal (MM) fusion strategies for text and vision features. Using FiLM layers leads to maximum retention of pre-trained VLM knowledge.}
\label{tab:app-ablation-4}
\end{table}

\noindent \textbf{Different text encoders: }We replaced the CLIP text encoder with the BERT encoder (bert-base-uncased) to understand the advantages of using a joint embedding space for images and text. BERT is pre-trained on large text datasets from the real world and provides high-quality features for referring queries. Results indicate that using CLIP significantly improves our scores across the test sets, validating the importance of a joint embedding space for effective visual grounding (Table \ref{tab:app-ablation-5}). We use this version of our model for the next set of experiments on open vocabulary settings (Section 4.5).

\begin{table}[!h]
\setlength{\tabcolsep}{3pt}
\centering
 \begin{tabular}{c c c}
 \toprule[1pt]
 \textbf{Text Encoder} & \textbf{Test Seen IOU} & \textbf{Test Unseen IOU}\\ 
\midrule[1pt]
 BERT & 80.12 & 58.98 \\
  \midrule[0.5pt]
 CLIP-Text & 85.73 & 70.74 \\
 \bottomrule[1pt]
 \end{tabular}
\caption{Ablation on different text encoders for referring queries. Using CLIP-Text encoder benefits our architecture design and removes the need for full-finetuning.}
\label{tab:app-ablation-5}
\end{table}

\section{RoboRES Data Creation}
\label{app:robores-data-creation}

\begin{figure*}[!ht]
\centering

\begin{subfigure}[b]{0.45\textwidth}
    \includegraphics[width=\textwidth]{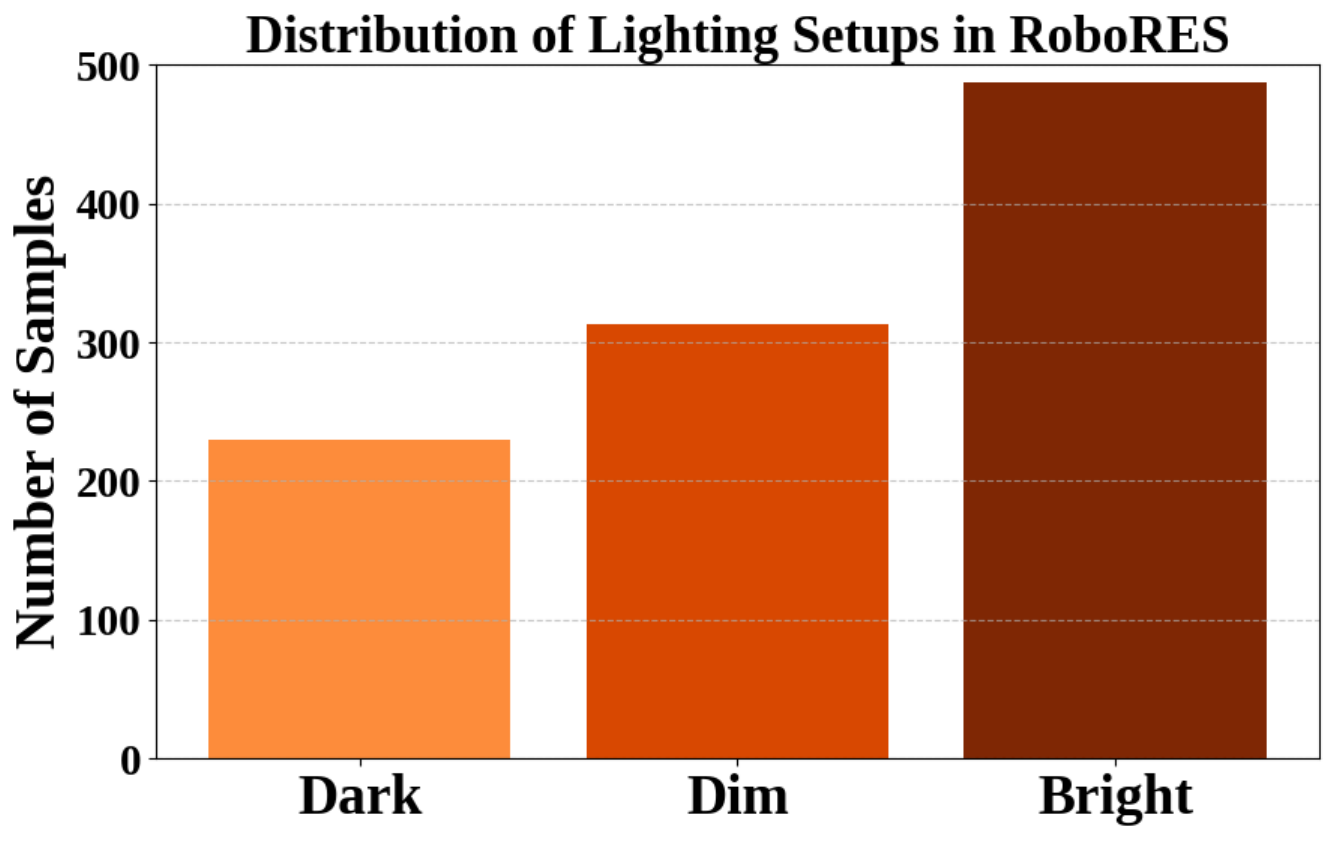}
    \label{fig:robores-a}
\end{subfigure}
\hfill 
\begin{subfigure}[b]{0.45\textwidth}
    \includegraphics[width=\textwidth]{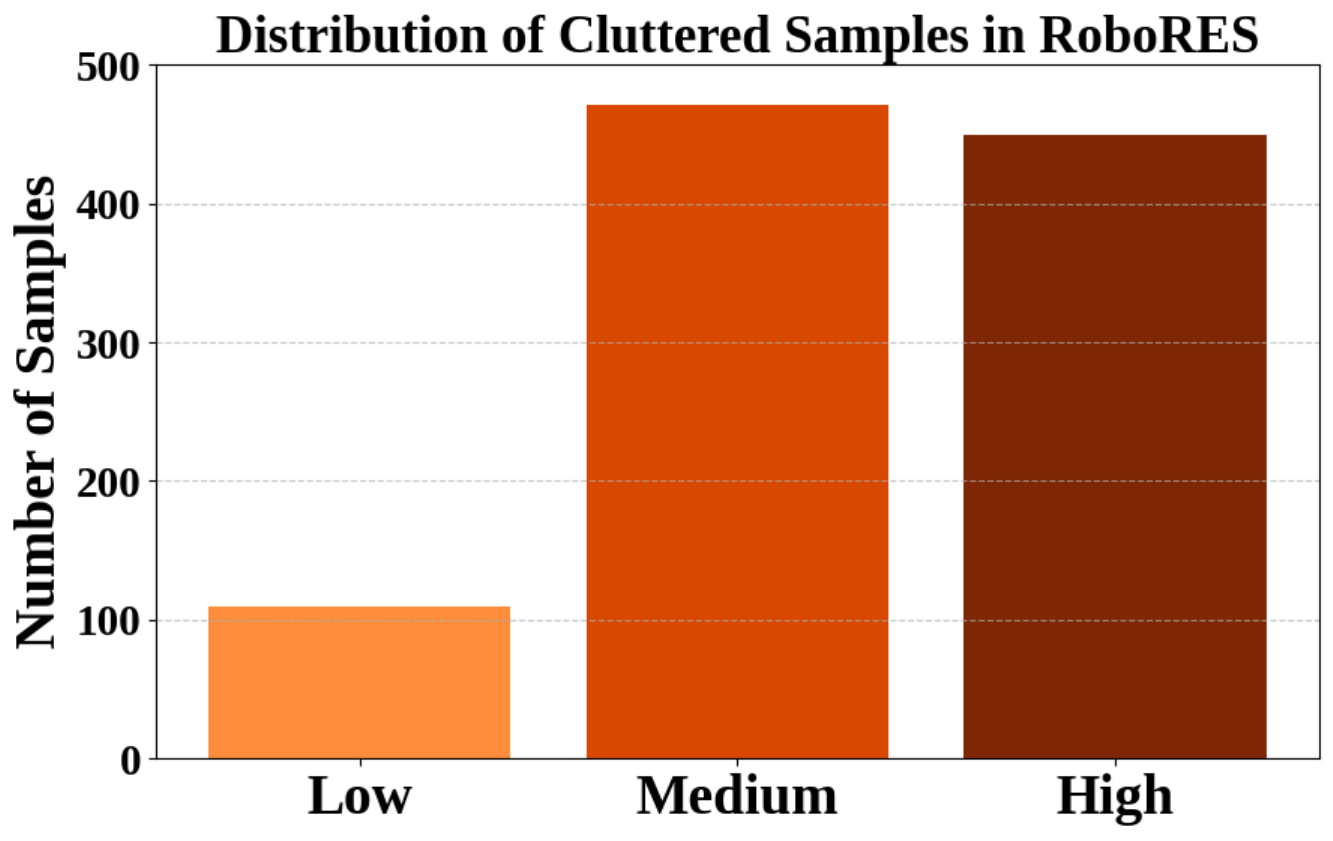}
    \label{fig:robores-b}
\end{subfigure}

\begin{subfigure}[b]{0.45\textwidth}
    \includegraphics[width=\textwidth]{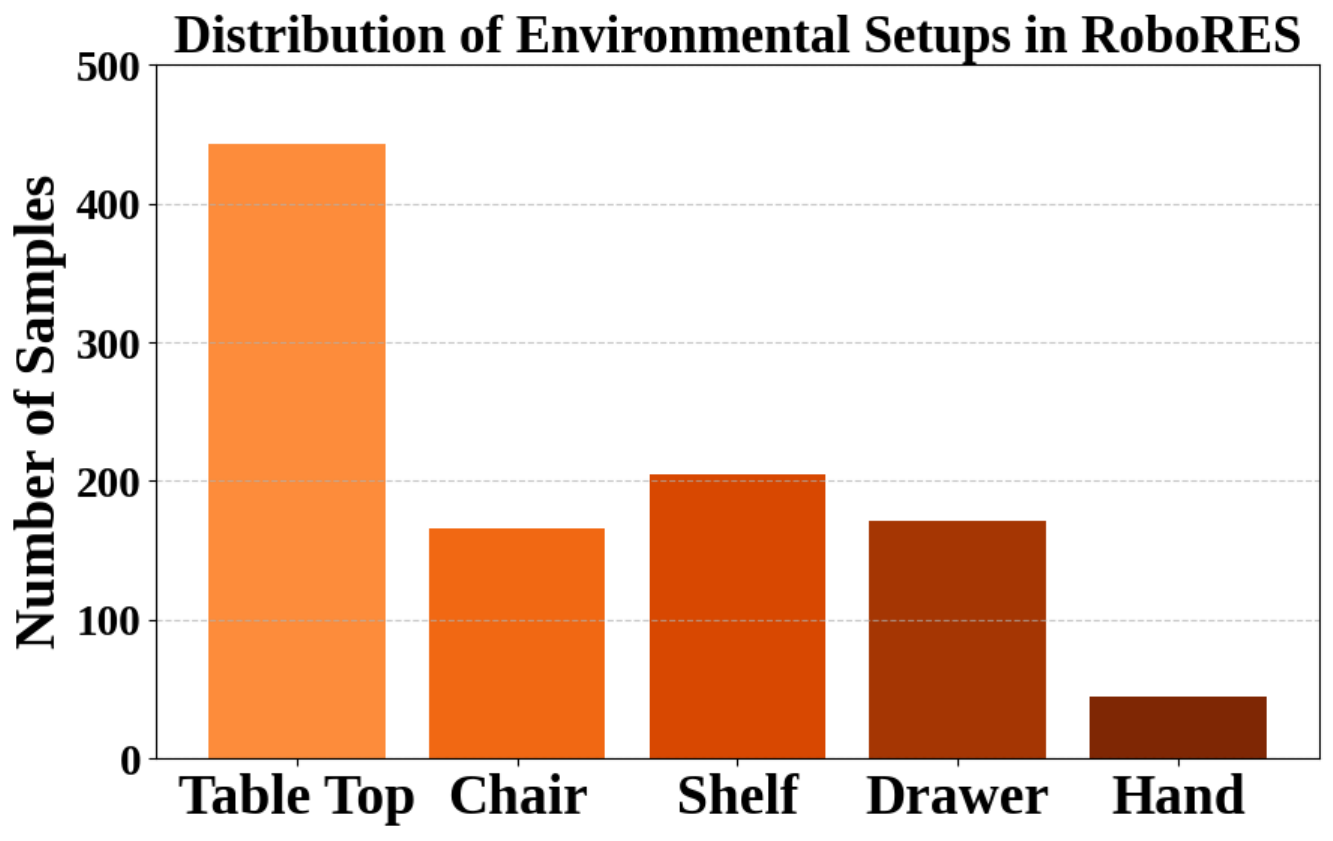}
    \label{fig:robores-c}
\end{subfigure}
\hfill
\begin{subfigure}[b]{0.45\textwidth}
    \includegraphics[width=\textwidth]{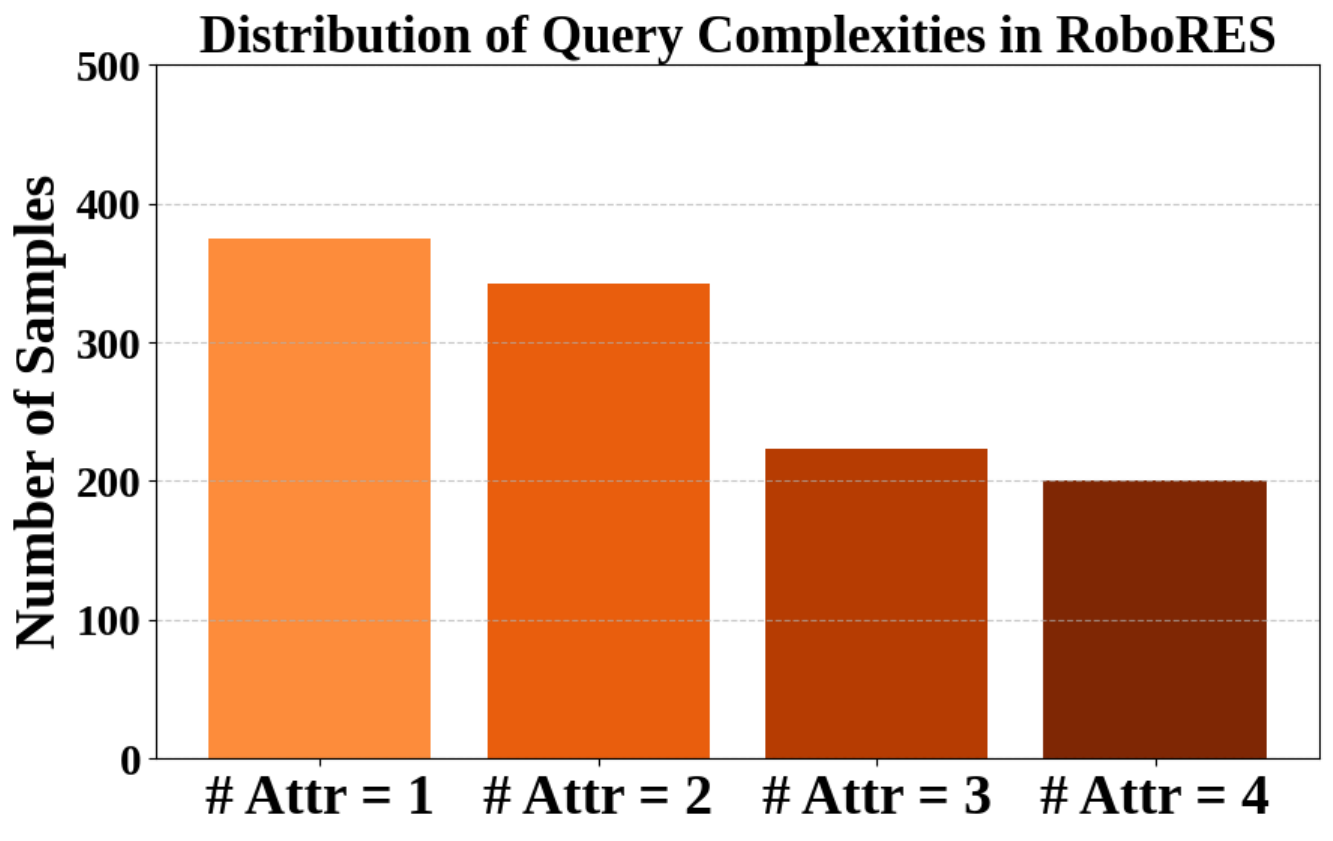}
    \label{fig:robores-d}
\end{subfigure}

\caption{Distribution of samples in RoboRES according to lighting conditions, clutter, environment, and query complexities. Here Attr denotes Attributes.}
\label{fig:robores-data-analysis}

\end{figure*}

To thoroughly benchmark our baselines in an open vocabulary setting with unseen samples, we created a new, complex test dataset to compare different visual grounding methods. The following steps outline the process of creating our corpus:

\begin{compactenum}
    \item \textbf{Selection of Objects and Environment:} 
    \begin{itemize}
        \item We selected a small set of graspable objects from day-to-day items.
        \item Considering the wide range of environments where grasping robots can be deployed, we decided on five setups for capturing images: Table Top, Chair, Multi-layered Shelf, Drawer, and Human Hand.
    \end{itemize}
    
    \item \textbf{Arrangement of Objects:}
    \begin{itemize}
        \item Objects were arranged with varying degrees of clutter:
        \begin{itemize}
            \item Low clutter (fewer than 4 well-spaced objects)
            \item Medium clutter (more than 4 closely spaced objects)
            \item High clutter (occluded objects present).
        \end{itemize}
        \item We also varied the lighting conditions: dark, dim, and bright, to ensure a holistic evaluation.
    \end{itemize}
    
    \item \textbf{Data Capture:}
    \begin{itemize}
        \item The environment and objects were set up, and images were captured using the RealSense D455 camera attached to our Franka Research 3 robotic arm.
        \item A total of 120 scenes were created with the given set of objects, varying clutter, lighting and background setup.
        \item We found that using the gripper camera was not necessary as similar results were obtained with a simple iPhone camera.
    \end{itemize}
    
    \item \textbf{Mask Generation and Verification:}
    \begin{itemize}
        \item The captured images were processed through the SAM model to generate candidate masks of all objects.
        \item Since not all masks corresponded to real objects, they were manually verified for accuracy.
    \end{itemize}
    
    \item \textbf{Crowd-Sourced Text Generation:}
    \begin{itemize}
        \item For each segmented mask, we crowd-sourced the generation of referring text among a group of 12 people.
        \item They were instructed to use minimal referring attributes to describe the object in the mask, but were encouraged to use as many attributes as necessary to uniquely identify the correct object in case of duplicates.
    \end{itemize}
    
    \item \textbf{Final Dataset:}
    \begin{itemize}
        \item Our final dataset consists of 1160 tuples of (RGB image, Mask, Text).
        \item Although this is not a very large corpus, our annotation process can be extended to scale the dataset as required.
        \item The distribution of the corpus across categories is provided in Figure \ref{fig:robores-data-analysis}. This corpus is used for our open-vocabulary experiments in Section 4.5.
    \end{itemize}
\end{compactenum}

\section{Real world experiments}
\label{app:real-world-exp}
We performed all real world grasping experiments on the Franka Research 3 robotic arm. The experimental setup involved using five common object categories: \textit{Fruit}, \textit{Soda Can}, \textit{Food Container}, \textit{Spray Bottle}, and \textit{Hardware}. For each category, there were three levels with an increasing number of distractors: Level 1 has one instance per object category, Level 2 has two instances per category, and Level 3 has three instances per category. For example, in Level 1, there are  5 objects, one from each category, with no distractors. In Level 2, there are 10 (5 $\times$ 2) objects, where each category has one target object and one distractor. In Level 3, there are 15 (5 $\times$ 3) objects, where each category has two distractors for the target object. This setup resulted in a total of 15 scenes, each designed to evaluate the model's ability to identify and grasp the target object in the presence of other items.

\subsection{Minimal Referring Query (MRQ)} MRQ refers to the query with the least number of attributes required to uniquely identify the object of interest within a given scene. This concept is particularly relevant in scenarios with multiple objects, where the goal is to pinpoint a specific object. For instance, if there are two apples in a scene, each apple can be uniquely identified with just one additional attribute (apart from ``apple" which is the object name). Possible MRQs for this scenario could be “Give me the apple on the right” or “Give me the smaller apple.” The MRQ is crucial for efficient and precise communication in robotic grasping tasks, minimizing query complexity while ensuring accurate object identification. In our experiments, if an MRQ results in the correct segmentation mask, we score that scene with 100 SA, as our model does not require redundant attributes to identify the target object accurately.

\begin{table*}[h!]
\centering
\begin{tabular}{>{\centering\arraybackslash}m{4.25cm} >{\centering\arraybackslash}m{4.25cm} >{\centering\arraybackslash}m{4cm}}
\multicolumn{3}{c}{\textbf{Grab the smaller red apple}}\\
\hline
\tiny \textbf{RGB Image} & \tiny \textbf{Predicted Segmentation} & \tiny  \textbf{Grasp Pose} \\
\includegraphics[width=\linewidth]{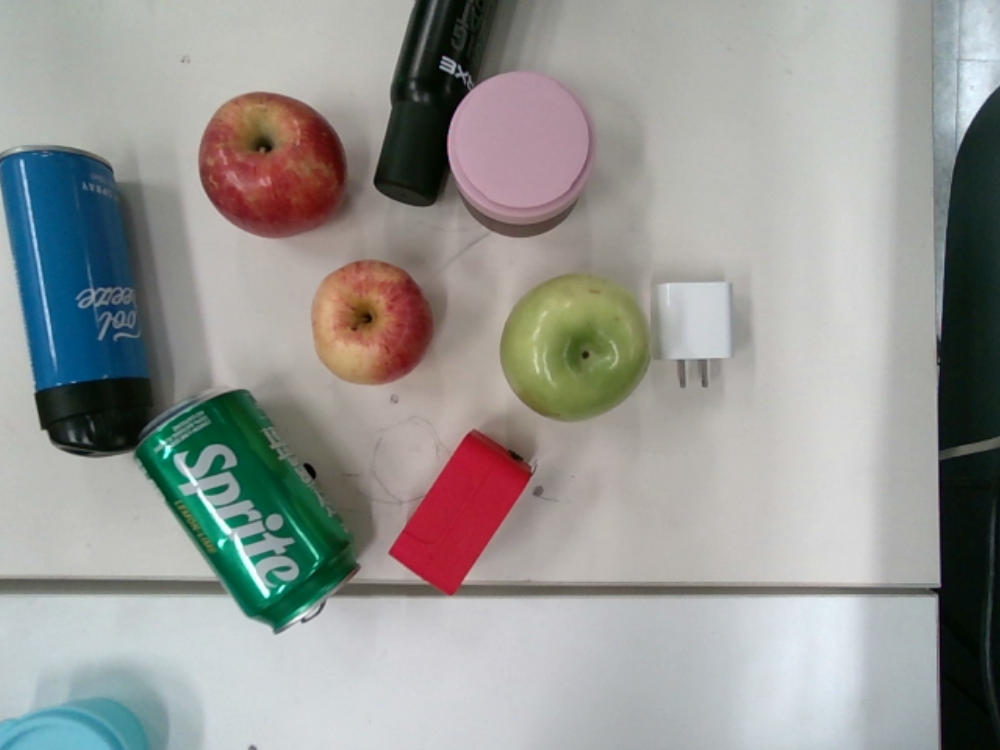} & \includegraphics[width=\linewidth]{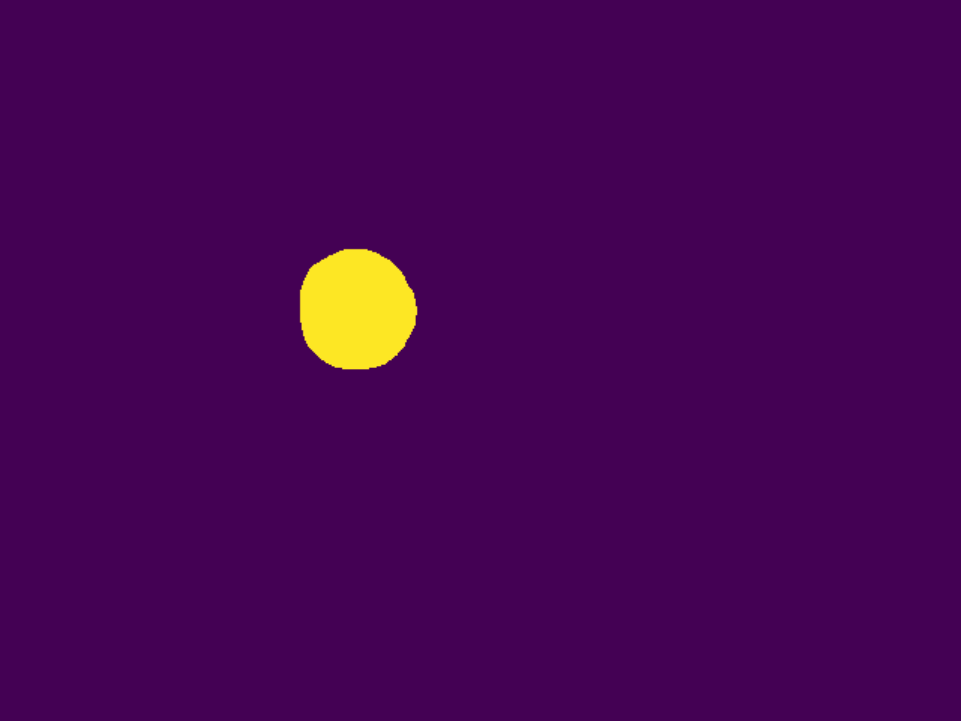} & \includegraphics[width=\linewidth]{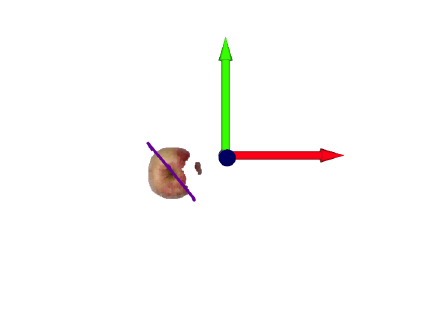} \\

\multicolumn{3}{c}{\textbf{Can you grab the blue soda can on the right?}}  \\
\hline
\tiny \textbf{RGB Image} & \tiny \textbf{Predicted Segmentation} & \tiny  \textbf{Grasp Pose} \\
\includegraphics[width=\linewidth]{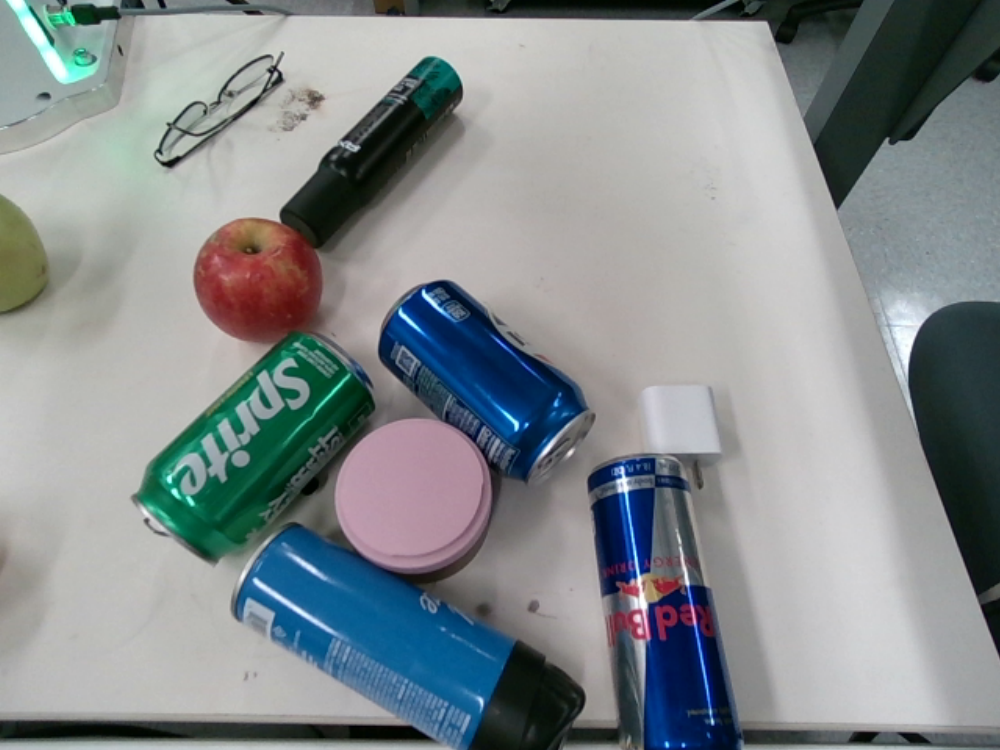} & \includegraphics[width=\linewidth]{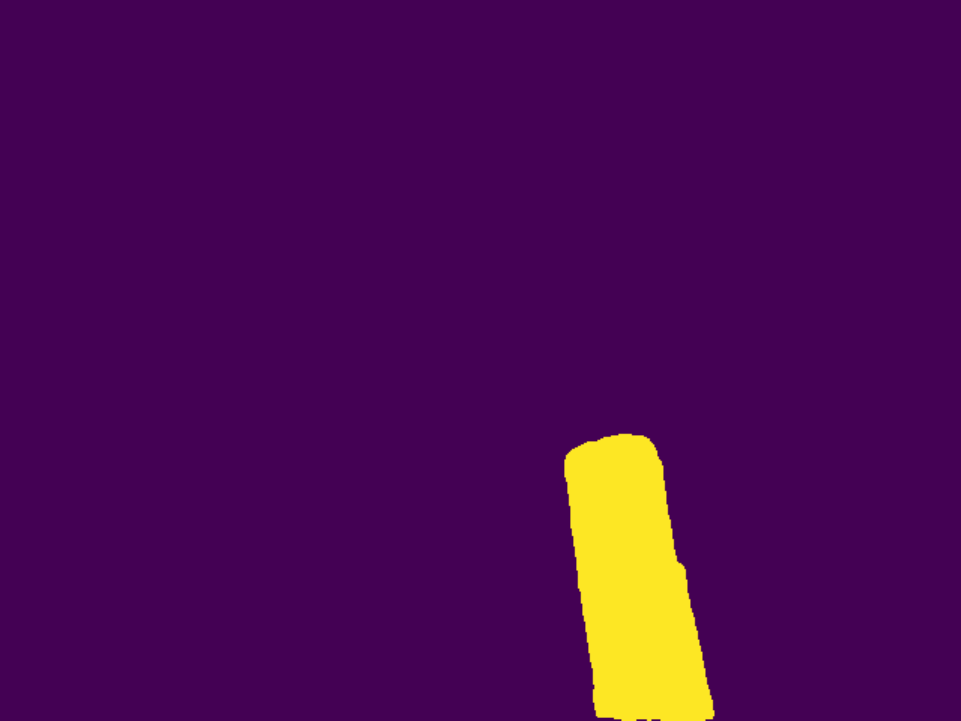} & \includegraphics[width=\linewidth]{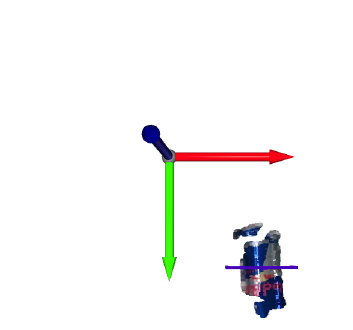} \\

\multicolumn{3}{c}{\textbf{Where is the blue smaller circular storage container?}} \\
\hline
\tiny \textbf{RGB Image} & \tiny \textbf{Predicted Segmentation} & \tiny  \textbf{Grasp Pose} \\
\includegraphics[width=\linewidth]{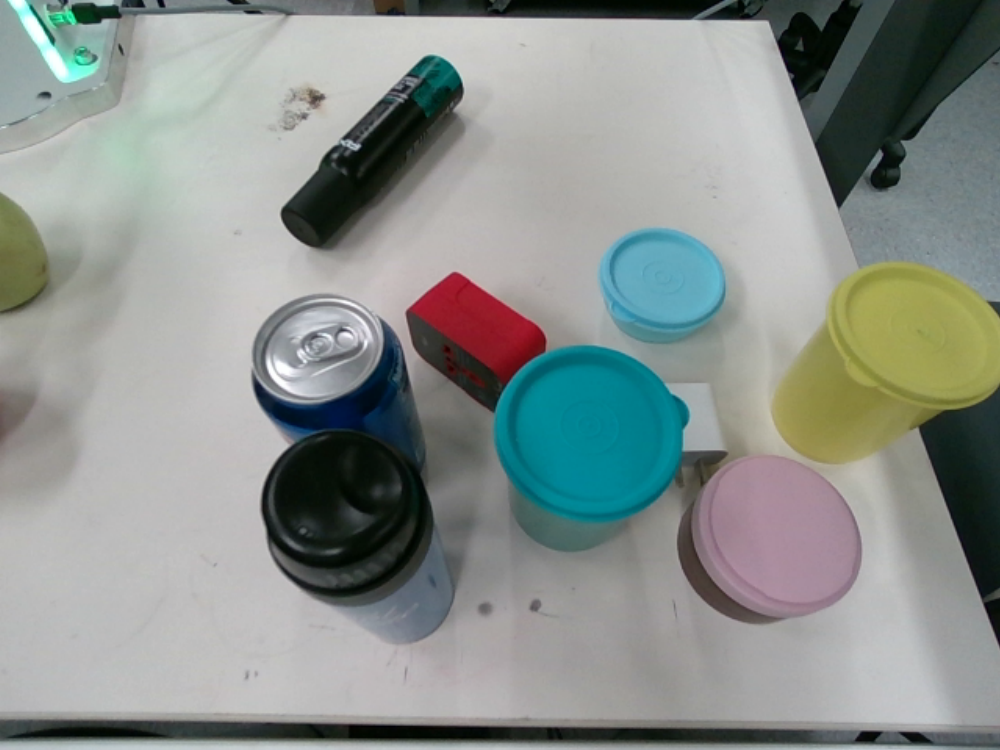} & \includegraphics[width=\linewidth]{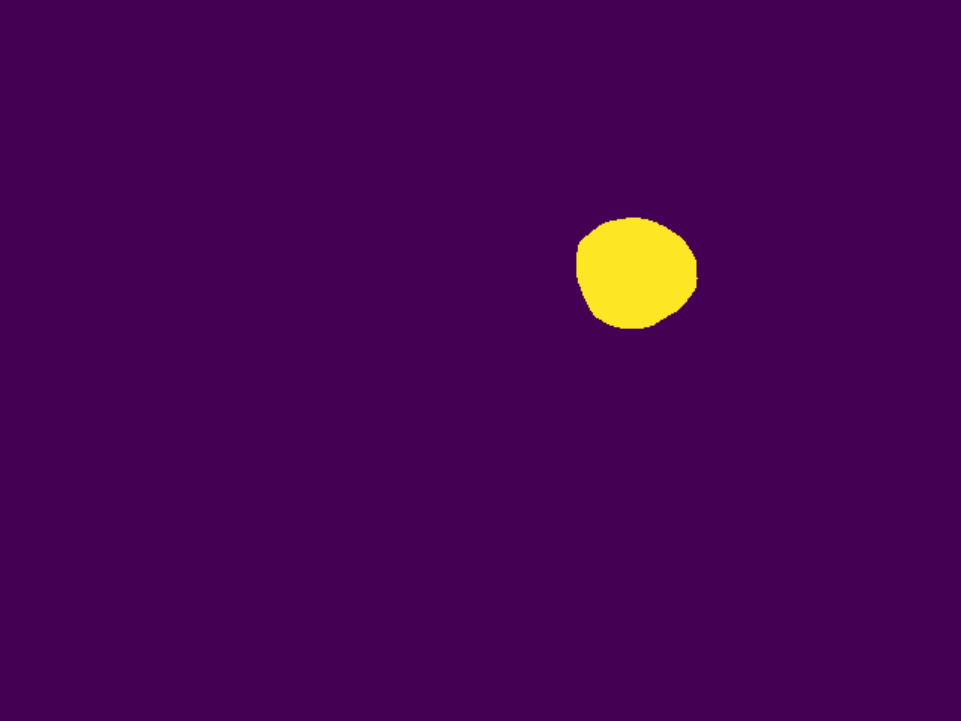} & \includegraphics[width=\linewidth]{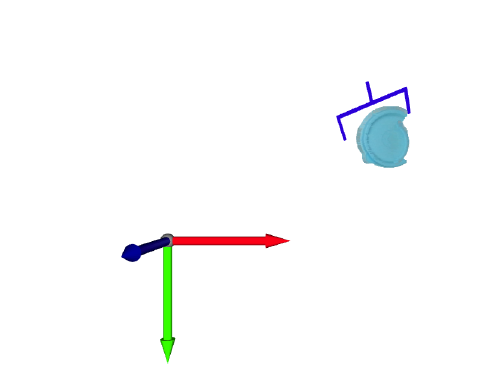} \\

\multicolumn{3}{c}{\textbf{Give me the black spray bottle on the left}} \\
\hline
\tiny \textbf{RGB Image} & \tiny \textbf{Predicted Segmentation} & \tiny  \textbf{Grasp Pose} \\
\includegraphics[width=\linewidth]{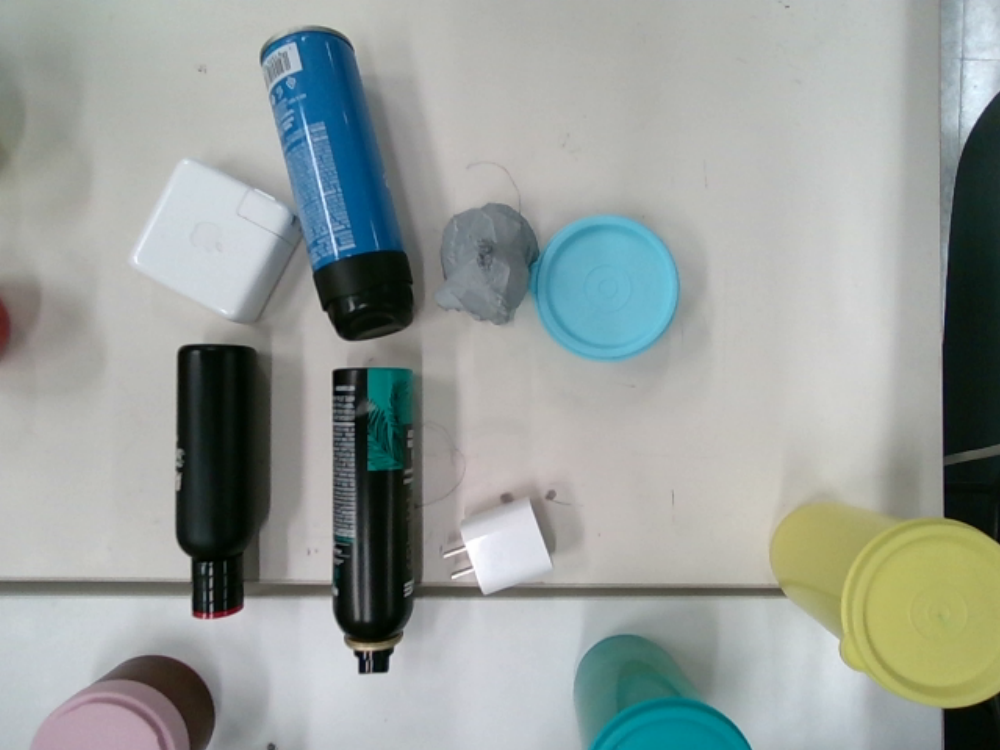} & \includegraphics[width=\linewidth]{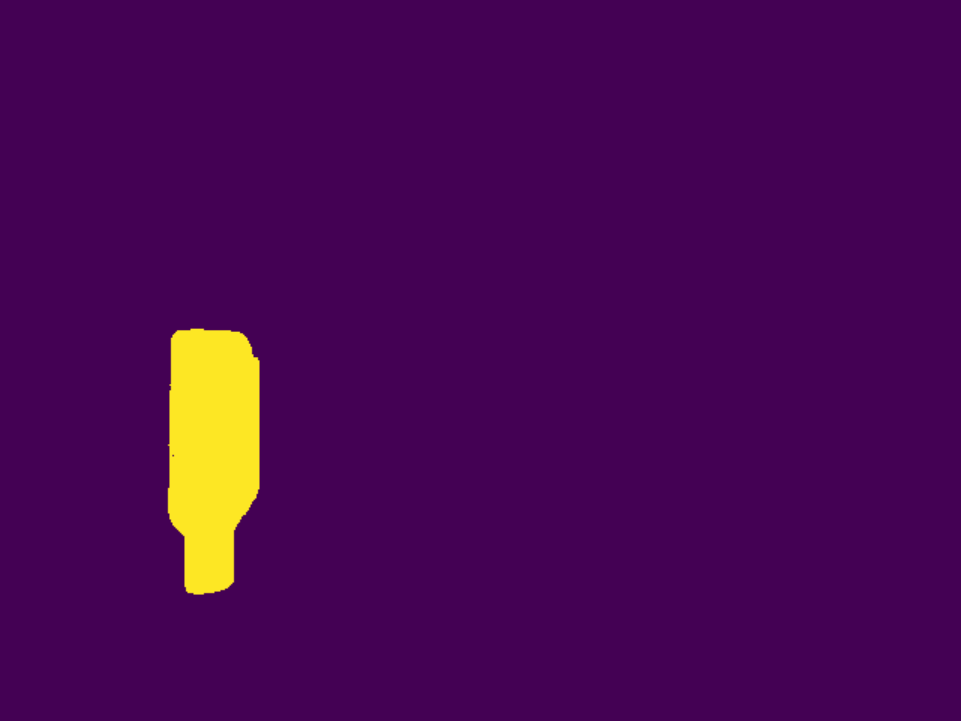} & \includegraphics[width=\linewidth]{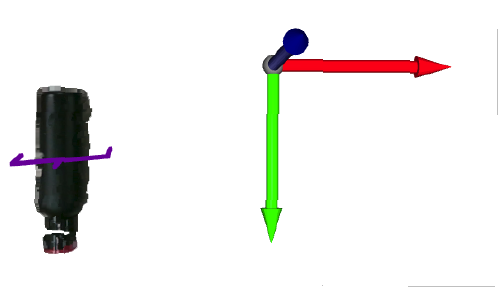} \\

\multicolumn{3}{c}{\textbf{Pick the smaller white charger adapter}}\\
\hline
\tiny \textbf{RGB Image} & \tiny \textbf{Predicted Segmentation} & \tiny  \textbf{Grasp Pose} \\
\includegraphics[width=\linewidth]{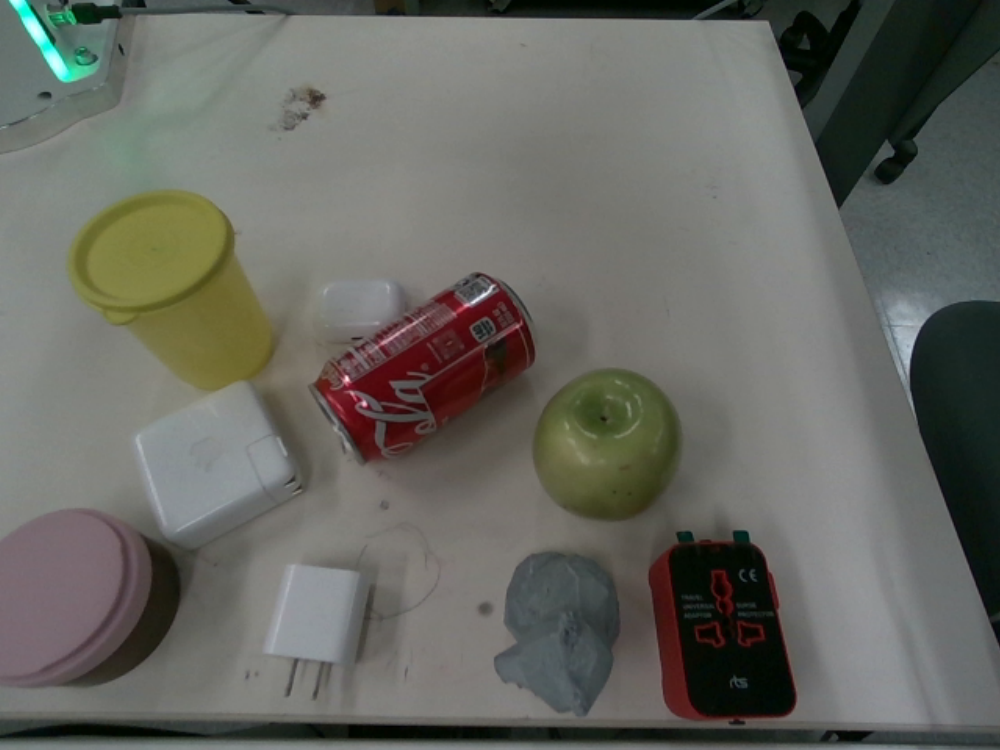} & \includegraphics[width=\linewidth]{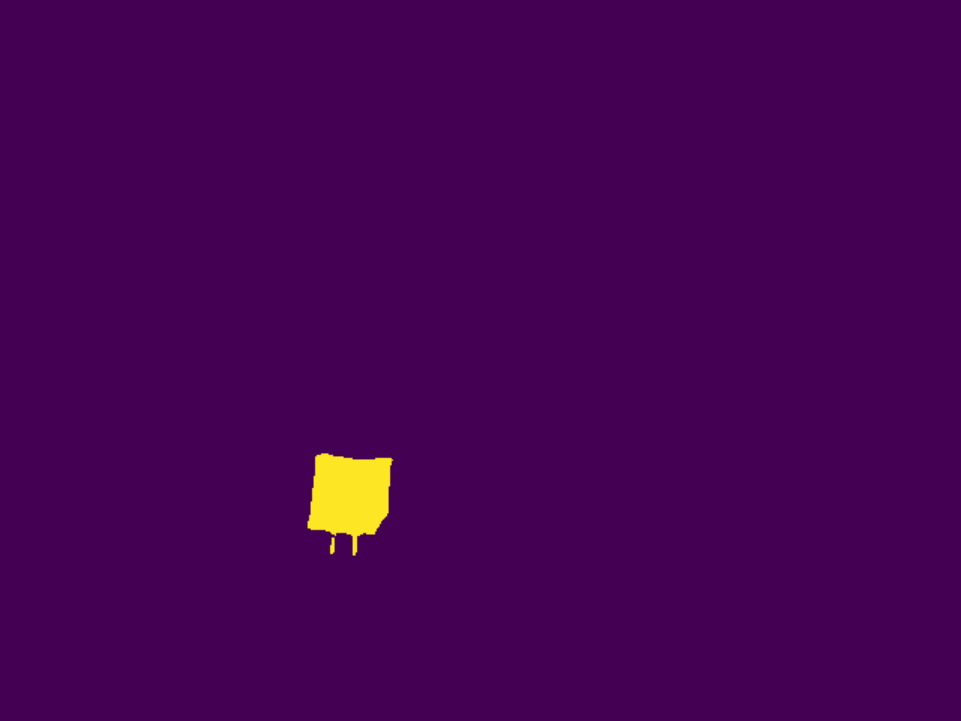} & \includegraphics[width=\linewidth]{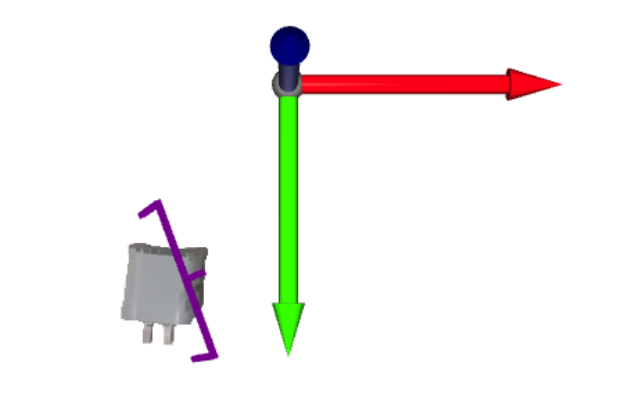} \\

\end{tabular}

\caption{Examples of real-world trials for Referring Grasp Synthesis. Our proposed approach generates segmentation masks from input RGB images and referring text queries. Segmented RGB-D images are used by AnyGrasp to output grasp pose parameters. }
\label{tab:experiment}
\end{table*}

\subsection{Implementing Referring Grasp Synthesis }The trials involved capturing RGB-D images with the robot, which provided both RGB (color) and depth information. These images, along with corresponding text queries, were input into the visual grounding model. This model segmented the referred object, producing the masked depth and RGB images, which were then processed by the AnyGrasp model to predict the grasp pose. The predicted 7 DOF grasp pose was subsequently executed by the robot. Table \ref{tab:experiment} showcases the predicted masks generated by GrSAM+HiFi-CS and grasp pose visualizations from AnyGrasp, with each image annotated with the corresponding input language query, displayed above each image. The examples include scenarios with multiple distractors to illustrate the model's robustness in complex environments. 

\begin{table*}[h!]
\centering
\begin{tabular}{>{\centering\arraybackslash}m{4cm} >{\centering\arraybackslash}m{4.25cm} >{\centering\arraybackslash}m{4.25cm}}
    \toprule[1pt]
    \textbf{Language Query} & \textbf{RGB Image} & \textbf{Predicted Segmentation} \\
    \midrule[0.5pt]
    \makecell[c]{Can you grab the larger blue \\ circular food container?} & \includegraphics[width=\linewidth]{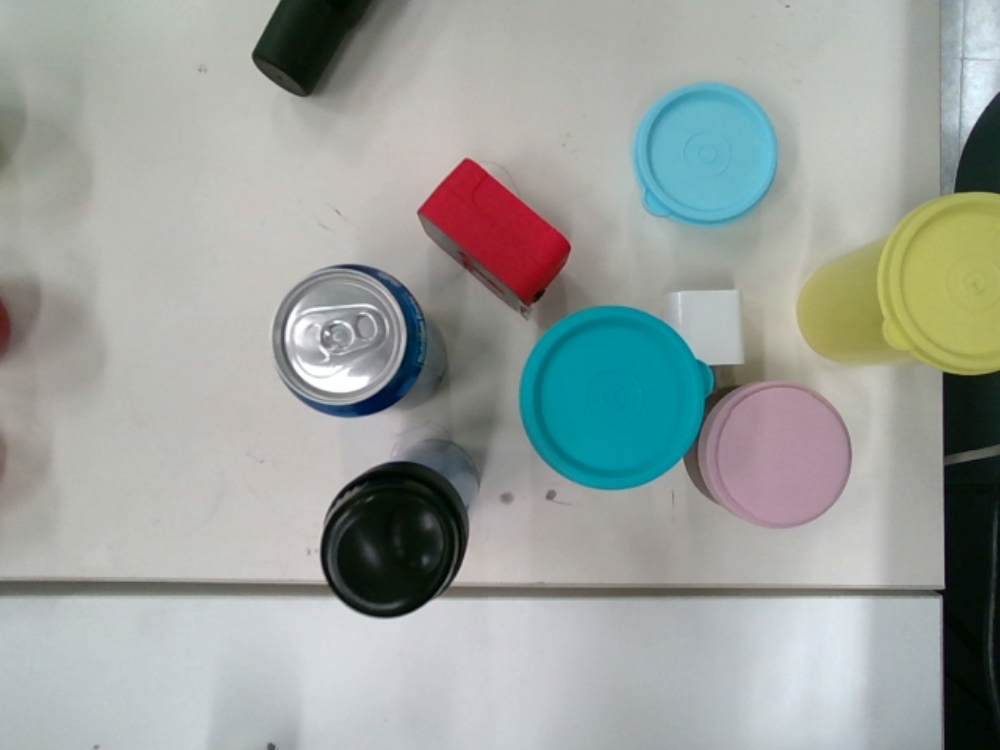} & \includegraphics[width=\linewidth]{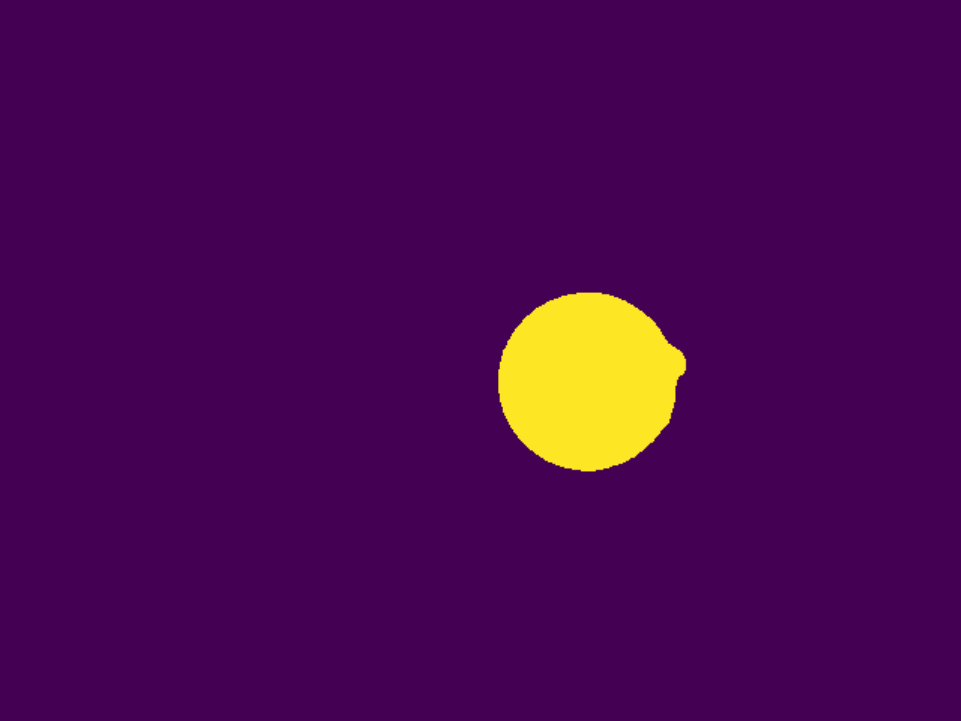} \\
    \hline
    \makecell[c]{Can you grab the larger blue \\ circular food container?} & \includegraphics[width=\linewidth]{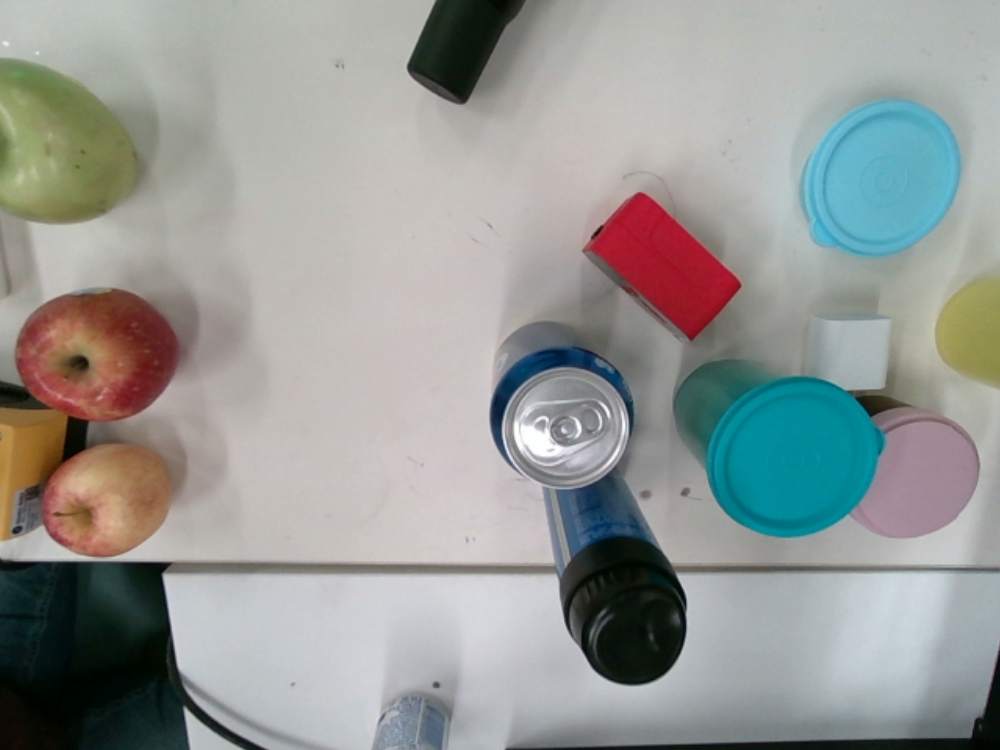} & \includegraphics[width=\linewidth]{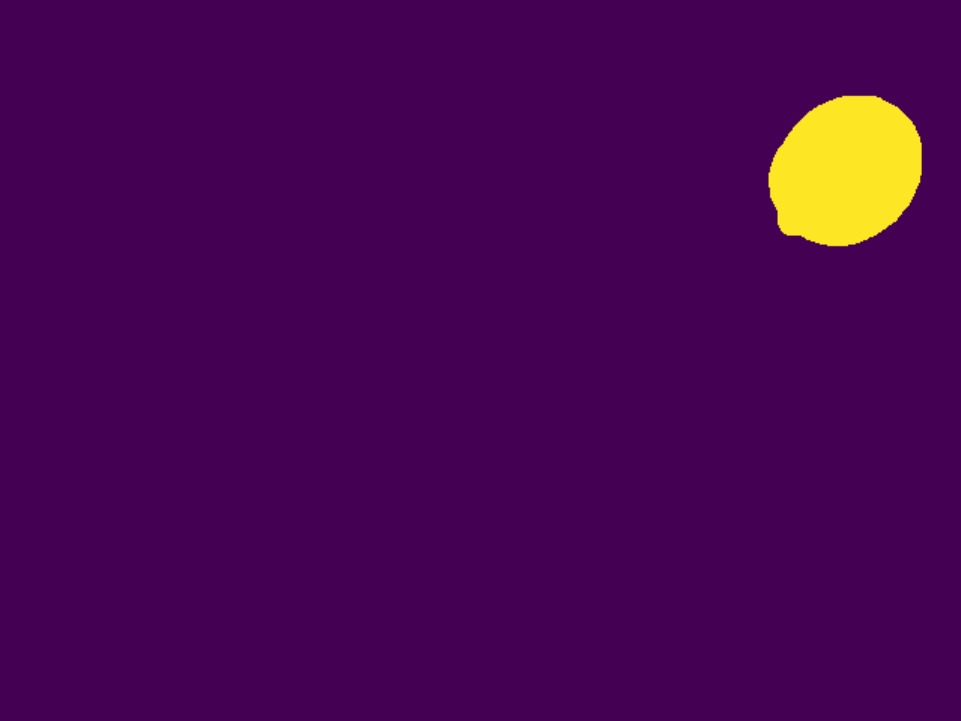} \\
    \midrule[0.5pt]
        \makecell[c]{Grab the blue soda can \\ on the bottom?} & \includegraphics[width=\linewidth]{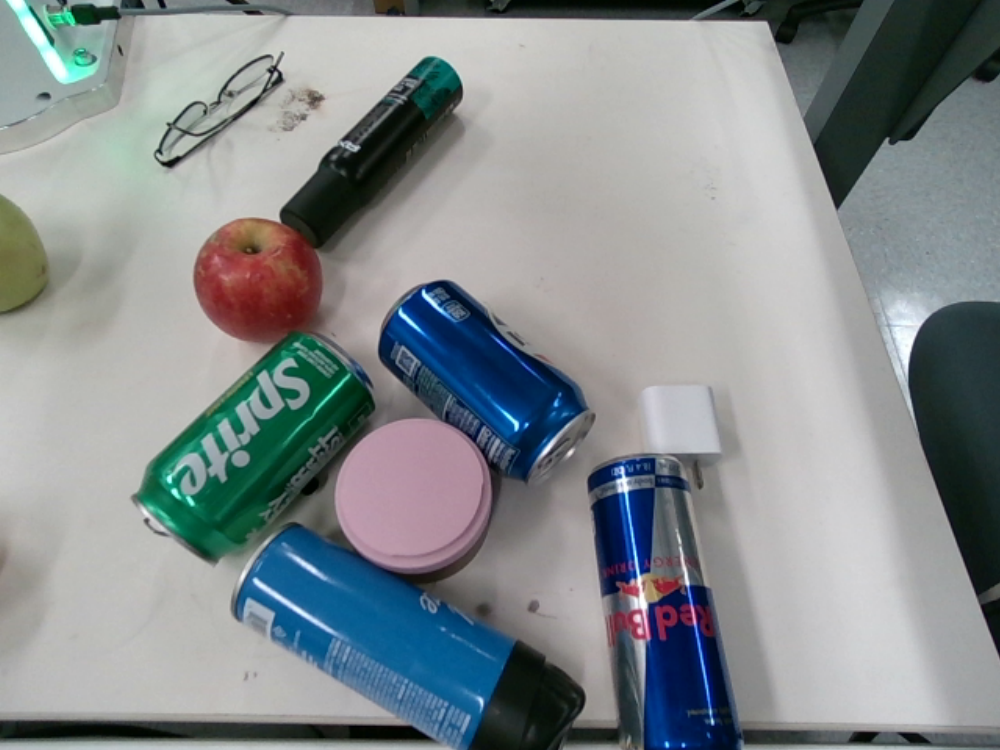} & \includegraphics[width=\linewidth]{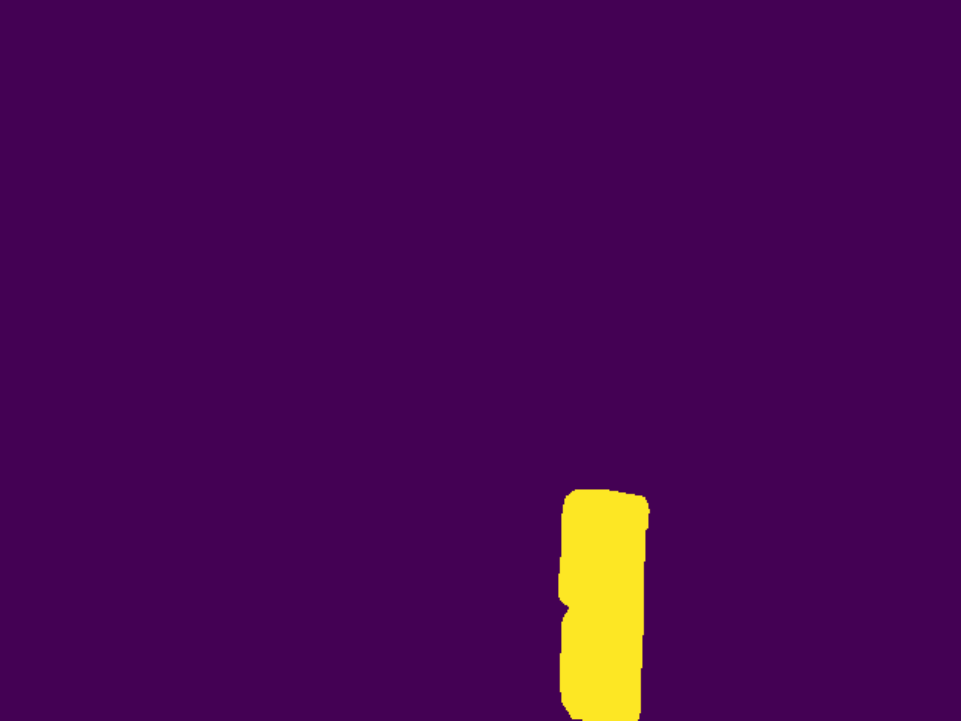} \\
        \hline
        \makecell[c]{Grab the blue soda \\ on the bottom?} & \includegraphics[width=\linewidth]{figures/3_rgb_1.pdf} & \includegraphics[width=\linewidth]{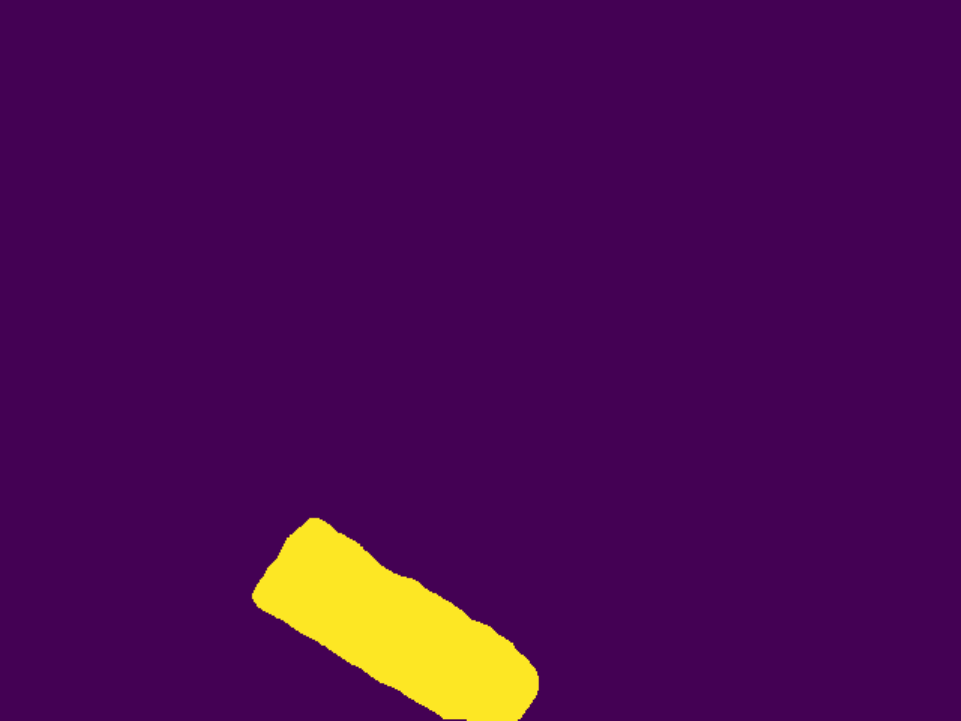} \\
        \bottomrule[1pt]
\end{tabular}
\vspace{5pt}
\caption{Instances of inaccurate predictions by GrSAM+HiFi-CS due to varying camera perspectives and minor changes in referring query}
\label{tab:experiment2_1}
\end{table*}

\subsection{Examining Segmentation Failure Cases }Table \ref{tab:experiment2_1} highlights instances where our visual grounding model failed. In the first two rows, we use the same query, ``Can you grab the larger blue circular food container?" for the images of the scene captured from different viewpoints. The model incorrectly identifies the smaller blue circular food container as the larger one in the second view. This discrepancy is due to the HiFi-CS model producing 2D segmentations, leading to perspective-dependent errors. In the top view, the blue container near the bottom looks bigger in perspective and our model correctly identifies this. However, in the second view, the blue container on the top looks larger than the blue container at the bottom due to a change in perspective. Therefore, the model misidentifies the smaller one as the larger one. Since the current approach does not utilize any 3D information, such segmentation inconsistencies may arise. Implementing 3D segmentation could mitigate these issues by providing more accurate, perspective-independent segmentations. 

In the next two rows, we use the same image and input different queries: ``Grab the blue soda can on the bottom?" and ``Grab the blue soda on the bottom?" In this case, the model wrongly predicts the blue spray bottle on the bottom as the blue soda when the word ``can'' is omitted from the query. This example illustrates the model's sensitivity to specific object names. A minor change in the query can lead to incorrect predictions, highlighting the importance of precise language. To achieve more accurate results, it is necessary to include additional attributes in the queries. This would help the model to better distinguish between objects, reducing the likelihood of mispredictions due to subtle differences in phrasing.

\subsection{Analysing Grasping Errors}

We highlight the problems with using only one camera for our real world experiments in this section. The partial point cloud constructed using this camera works well for solid shapes like hardware adapters, as the solid edges of these objects are clearly represented in a top view. However, for curved objects like apples, soda cans, and spray bottles, the top view point clouds sometimes fail to accurately depict their exact curvature, resulting in a small offset during grasp pose execution. The widths of the soda can (6.6 cm) and food container (7.6 cm) are approximately the same as the maximum width of the gripper (8 cm), so even a tiny offset can lead to failure. Since the diameter of the spray bottle is smaller (5.3 cm), the offset does not cause an error. Another reason for the low grasping success rate is over-prediction in the visual grounding stage. Baselines such as GroundedSAM sometimes segment more than one object instance of the same/similar category, and although the segmentation mask contains the referred object, the best grasp pose might be executed on another object. Using a combination of our fine-tuned model, HiFi-CS, with an open-set detector like GroundedSAM helps prevent such over-predictions, improving grasping accuracy, as shown in Table 5 of the main paper. Adding more cameras would help construct a better point cloud and aid in grasp pose accuracy. The novelty of our approach lies in the visual grounding model, and we demonstrate that our combined approach leads to overall improvements, which can greatly benefit tasks like Referring Grasp Synthesis. We provide overall conclusions and directions for future work in Section 6.